\documentclass[11pt]{article}

\usepackage[preprint]{acl}

\usepackage{times}
\usepackage{latexsym}

\usepackage[T1]{fontenc}

\usepackage[utf8]{inputenc}

\usepackage{microtype}

\usepackage{inconsolata}

\usepackage{graphicx}
\usepackage{acl}
\usepackage{diagbox}

\usepackage{times}
\usepackage{colortbl} 
\usepackage{booktabs} 
\usepackage{tabularx}
\usepackage{fontawesome}

\usepackage[table]{xcolor} 

\usepackage{amsfonts}
\usepackage{xcolor}
\usepackage{adjustbox}
\usepackage{listings}

\usepackage{latexsym}
\usepackage{enumitem}
\usepackage{floatrow} 
\usepackage{booktabs}
\usepackage{longtable} 
\usepackage{colortbl} 
\usepackage{adjustbox} 
\usepackage{array} 
\usepackage{amsmath}  
\usepackage{cleveref}
\usepackage{adjustbox}
\usepackage{xspace}
\usepackage{graphicx}    
\usepackage{caption}     
\usepackage{subcaption}  
\usepackage{afterpage}   
\usepackage{longtable}
\usepackage{makecell}
\usepackage{natbib}
\usepackage{multirow}
\usepackage{todonotes}  
\newcommand{\dataset}{\texttt{StyloBench}\xspace}
\newcommand{\method}{\texttt{StyloCheck}\xspace}
\usepackage[table]{xcolor}
\usepackage{xcolor}
\usepackage{colortbl}
\renewcommand{\textcolor}[2]{#2} 
\title{When Personalization Tricks Detectors:\\ The Feature-Inversion Trap in Machine-Generated Text Detection}

\author{
\textbf{Lang Gao}$^{1}$,~~\textbf{Xuhui Li}$^{1}$,~~\textbf{Chenxi Wang}$^{1}$,~~\textbf{Mingzhe Li}$^{2}$,~~\textbf{Wei Liu}$^{3}$,\\
~~\textbf{Zirui Song}$^{1}$,~~\textbf{Jinghui Zhang}$^{1}$,~~\textbf{Rui Yan}$^{4}$,~~\textbf{Preslav Nakov}$^{1}$,~~\textbf{Xiuying Chen}$^{1}$\thanks{Corresponding author.}\\
\\
$^{1}$MBZUAI~~
$^{2}$ByteDance\\
$^{3}$National University of Singapore~~
$^{4}$Wuhan University\\
\texttt{\{Lang.Gao, Preslav.Nakov, Xiuying.Chen\}@mbzuai.ac.ae}\\
}

\begin{document}
\maketitle
\begin{abstract}
As large language models (LLMs) increasingly imitate personal writing styles, personalization has become a key challenge for machine-generated text (MGT) detection. Yet personalized MGT detection remains largely underexplored.
In this work, we introduce \dataset, the first benchmark for evaluating detector robustness under personalization, built from literary and blog texts paired with their LLM-generated imitations.
Experiments across diverse detectors show pronounced performance instability under personalization, with frequent inversions relative to general-domain behavior.
\textcolor{blue}{
To better understand this limitation, we conduct an in-depth analysis and attribute it to a \textit{feature-inversion trap}, i.e., features that are effective for separating human-written text (HWT) from MGT in general flip their effect in personalized contexts, ultimately misleading detectors.
Motivated by this, we propose \method, a diagnostic framework for predicting detector robustness under personalization. \method identifies the inverted features and quantifies detector dependence using perturbed texts pronounced in the features.}
In our experiments, \method predicts both the direction and magnitude of cross-domain performance shifts with an 85\% correlation to actual outcomes.
We hope this work will raise awareness of the structural risks introduced by personalization and motivate more robust approaches to personalized MGT detection.
\faGithub~\href{https://github.com/mbzuai-nlp/Personalized_MGT_Detect}{Github.}

\end{abstract}

\section{Introduction}


\begin{figure}[htbp] 
\centering
\includegraphics[width=\columnwidth]{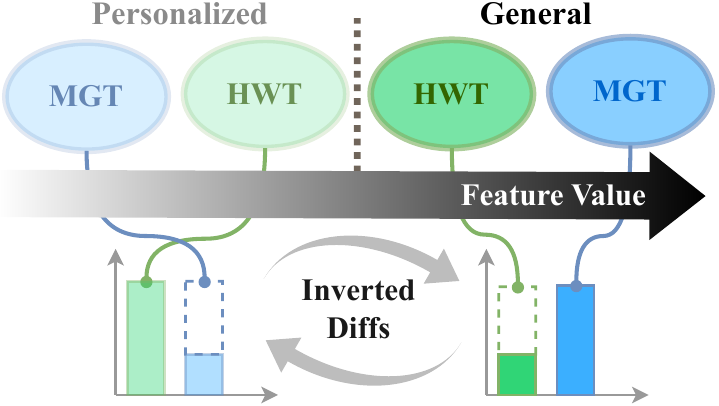} 
\caption{
Illustration of the feature-inversion trap. The feature values of HWT/MGT exhibit inverted differences across domains.
}
\label{fig:p1}
\end{figure}


Large Language Models (LLMs) have achieved strong text generation performance~\cite{huang2025trustworthiness}, with increasing capability of mimicking personalized language styles in tasks such as news writing, style imitation, and story generation~\cite{tu2024charactereval,wang2025beyond}, across diverse domains, settings, and increasingly complex real-world applications.

However, these capabilities also raise security and ethical concerns, as LLMs may generate fake news and misinformation~\cite{tian2025symbolic}.
Moreover, style imitation can be misused, for instance, by impersonating public figures or creating fake work~\cite{herbold2024largelanguagemodelsimpersonate}.
These risks make machine-generated text (MGT) detection increasingly important and urgent in real-world deployment scenarios across platforms, media ecosystems, and high-stakes communication environments globally.
Although existing studies have made progress in general-domain detection, it remains unclear how well they perform in personalized domains with diverse stylistic variations and user-specific nuances, preferences, and contextual dependencies.
Hence, we introduce \dataset, the first benchmark for MGT detection in personalized settings.
\dataset covers two sub-scenarios: Literary works and Blog posts, each paired with LLM-generated imitations.
Interestingly, experiments on \dataset show that personalization can noticeably degrade detector performance. Moreover, on the Literary subset, many detectors exhibit \textit{prediction inversion}, where predictions tend to shift opposite to the expected direction, suggesting weakened or even flipped discriminative cues under highly personalized text.


In order to explore why detectors fluctuate under personalization, we train a domain classifier on \textcolor{blue}{human-written texts (HWT)} from general and personalized domains, and test it on both HWT/MGT in the two domains to see, across domains, whether domain features fluctuate similarly and consistently across different evaluation settings.
The classifier exhibits a clear inversion: In the general domain, MGT is predicted to be more general'' than HWT; however, in the personalized domain, MGT is instead classified as more personalized'' than HWT.


This has given rise to our \textit{feature-inversion trap} hypothesis, namely that features that are effective for separating HWT from MGT in general
flip their effect in personalized contexts, ultimately misleading detectors.
In order to identify where this inversion is most pronounced, we formalize the search for the strongest inversion as a Rayleigh quotient problem~\cite{dong2024rayleigh} and exploit its extremal property to obtain the \textit{inverted feature direction}.
Feature directions derived from different datasets exhibit high cosine similarity, closely aligned with domain classifier weights and far from those of HWT/MGT classifiers.
Projecting samples onto these feature directions yields scalar \textit{feature values}, as shown in Figure~\ref{fig:p1}, which reveal a clear inversion effect: within each dataset, the HWT–MGT feature value difference is negatively correlated with detector performance.
This suggests that the ``feature-inversion trap'' is a stable, cross-domain phenomenon akin to stylistic differences, and that detector failures are partly driven by reliance on these inverted features.

Based on this finding, we propose \method, an effective approach to predict the detector's performance changes in personalized scenarios. 
Given a detector, \method evaluates it on probe datasets constructed with token-level perturbations that remove semantics, style, and basic HWT/MGT features while preserving inverted-feature differences.
The resulting performance reflects the detector's reliance on inverted features: higher performance on probe datasets indicates stronger reliance on these features.
We test seven detectors on 100 probe datasets and find that the Pearson correlation between \method's outputs and the actual cross-domain performance gaps exceeds 0.7 in 90\% of the cases, and consistently stays above 0.85.
This shows that \method reliably predicts both the direction and the magnitude of transfer performance changes, with higher reliability as the number of probe datasets increases.


\textcolor{blue}{
\begin{itemize}
\item We build \dataset, the first benchmark for MGT detection in personalized scenarios, and uncover drastic performance declines and even reversals in existing detectors.
\item We identify the \textit{Feature-Inversion Trap}, a systematic shift between general and stylized domains, and show that this phenomenon can fundamentally undermine detectors by inverting the features they rely on.
\item We propose \method, which estimates both the direction of change and the magnitude of performance variation of a detector under personalized scenarios. It serves as an early warning signal without requiring large-scale testing. The estimation shows high reliability and strong consistency with actual performance, with Pearson correlation exceeding 0.85.
\end{itemize}
}

\section{Related Work}

\subsection{MGT Detection}
Several benchmarks have been developed for evaluating the performance of HWT vs. MGT detection across different domains, generators, and languages, e.g., MGTBench~\cite{mgtbench0001SC0024}, \texttt{M4}~\cite{wang-etal-2024-m4}, \texttt{M4GT-Bench}~\cite{WangMISSTAM0AAH24} and RAID~\cite{DuganHTZLXIC24}.
Moreover, several domain-specific benchmarks, including WetBench~\cite{quaremba2025wetbenchbenchmarkdetectingtaskspecific} and MultiSocial~\cite{MackoKMS25}, have focused on specialized contexts such as Wikipedia and social media. However, none of them has looked into evaluating MGT detection in highly personalized or stylistically consistent text.

MGT detection methods fall into two main categories: training-based and training-free~\cite{xu2025trainingfree}. Training-based methods treat detection as supervised classification, usually by fine-tuning pretrained encoders such as RoBERTa~\citep{liu2019robertarobustlyoptimizedbert}, or by using improved frameworks and architectures~\citep{guo2024detective,tian2024multiscale,jiao-etal-2025-rangedetector}. Training-free methods rely on explicit textual or probabilistic cues, including geometric and probabilistic signals~\citep{bao2024fastdetectgpt,xu2025trainingfree}, token distributions~\citep{su2023detectllm}, topological features~\cite{tulchinskii2023intrinsic,wei2025shortphd}, and human-assistive indicators~\citep{gehrmann-etal-2019-gltr,russell-etal-2025-people}.
However, existing studies have not addressed or evaluated personalized or highly stylistically adaptive scenarios.

\subsection{Personalized LLM Generation}

Personalization of LLMs has recently become increasingly important~\cite{zhang2025personalization}. Personalization methods have developed into two main approaches: (\emph{i})~\emph{Prompt-based personalization} drives LLMs toward users' traits via personalized prompts~\cite{tseng-etal-2024-two}. Some work designed retrieval~\cite{mysore-etal-2024-pearl} and agent frameworks~\cite{zhang2025personaagentlargelanguagemodel} to achieve deeper imitation and more context-aware, user-aligned generation behaviors.
(\emph{ii})~\emph{Training-based personalization} adapts user traits via instruction tuning~\cite{wozniak2024personalized,liu-etal-2025-llms}, or through self-supervised learning for dynamic adaptation~\cite{mendoza2024adaptiveselfsupervisedlearningstrategies}. This often yields stronger and more persistent stylistic alignment across diverse tasks and long-form generation settings.
However, such personalization ability of LLMs raises concerns, including the possibility for political impersonation~\cite{herbold2024largelanguagemodelsimpersonate} and copyright infringement~\cite{zhang-etal-2025-llms,karamolegkou2023copyright}.
This underscores the need for personalized MGT detection. However, to the best of our knowledge, no prior work has systematically studied the problem of personalized MGT in realistic, large-scale evaluation scenarios.

\section{\dataset}
To investigate the performance of existing MGT detection methods in personalized scenarios, we create \dataset, the first benchmark for MGT detection under personalized conditions. This dataset has two subsets representing two scenarios: (\emph{i})~\texttt{Stylo-Literary}, simulating personalization in literary works, and (\emph{ii})~\texttt{Stylo-Blog}, simulating personalization in blogs. Table~\ref{tab:basic_info} gives some statistics about these datasets. 
\textcolor{blue}{We provide a detailed discussion about motivation for dataset scenario selection, model selection, and dataset scale and diversity in Appendix~\ref{app:disc-dataset}}.
\begin{table}[t]
\caption{Statistics about \dataset. CPT: Continuous Pretraining.}
\centering
\small
\begin{tabular}{@{}l c c@{}} 
\toprule
Subset & \texttt{Stylo-Literary} & \texttt{Stylo-Blog} \\
\midrule
Domain        & Article  & Blog \\
Generator Size   & $\leq 14$B   & $\geq70$B \\
Method        & CPT      & Prompting \\
Generators    & 3        & 4 \\
Subdomains  & 7        & 1 \\
Examples   & 21,000    & 4,000 \\
Sample Length & $\leq 512$ tokens & $\leq 512$ tokens \\
\bottomrule
\end{tabular}%
\label{tab:basic_info}
\end{table}

\subsection{Dataset Construction}
\subsubsection{\texttt{Stylo-Literary}} 
In the article scenario, HWTs consist of excerpts from literary works, while MGTs are generated by LLMs trained to learn and imitate the authors' styles.
Concretely, for HWT, we use data from the Gutenberg Book Corpus~\cite{e22010126}, an open-source collection of books grouped by authors. We selected seven authors: Jane Austen (J.A), Charles Dickens (C.D), Fyodor Dostoyevsky (F.D), Plato (P.L), Bernard Shaw (B.S), Jonathan Swift (J.S), and Mark Twain (M.T). These authors are well known for their distinctive styles and each has more than five long-form works. As some artifacts in the texts are not original content, but rather formatting or source information, we clean the texts to keep only the original content (see Appendix~\ref{app:stylo-liter} for more detail). Then, we split each author's texts into 512-token segments. 
For each author, we randomly select
1,000 segments as HWTs in the test set, and up to 3,000 additional segments as the training set.
For MGT, we apply \textcolor{blue}{Continuous Pretraining~\citep[CPT,][]{shi2024continual}}
 of LLMs on the training set to achieve deeper personalized imitation.
We train three LLMs in their base versions: Qwen3-4B~\cite{qwen3technicalreport}, Llama-3.1-8B~\cite{llama3}, and \textcolor{blue}{Phi-4 \citep[14B,][]{abdin2024phi4technicalreport}}.
We update only one LoRA~\cite{hu2022lora} layer to reduce the training cost and to speed up learning. After training, we take the first 30 tokens of each HWT test sample as the input and let the LLM continue the text. 
The selected hyperparameters and other generation details are in Appendix~\ref{app:llm-config}.

\subsubsection{\texttt{Stylo-Blog}} 
In the blog scenario, the HWTs come from Blog-1K~\footnote{\url{https://zenodo.org/records/7455623}}, a high-quality subset of the Blog Authorship Corpus~\cite{schler2006effects}. Blog-1K contains multiple posts grouped by 1,000 human authors. We further introduce the data source in Appendix~\ref{app:stylo-blog}.
We randomly select 1,000 posts, each truncated to a maximum length of 512 tokens, as HWT examples in the test set, and the corresponding MGTs are generated by LLMs. 
For each blog post, we apply a few-shot prompting template using 1--3 other posts by the same author as examples to guide the model in imitating the author's style.
The generator continues from the first 30 tokens of the given post, producing text with approximately the same length as the original. 


We use the following four popular large-scale LLMs as generators: GPT-4o, Claude-4-Sonnet (Claude-4), Claude-3.7-Sonnet-Latest (Claude-3.7), and Qwen2.5-72B, covering diverse architectures. The full generation details are in Appendix~\ref{app:stylo-blog} and \ref{app:llm-config}.


\begin{table*}[t]
\centering
\small
\setlength{\tabcolsep}{4pt}
\resizebox{\textwidth}{!}{
\begin{tabular}{lccccccccccc}
\toprule
 & \multicolumn{4}{c}{\texttt{M4}(General)} & \multicolumn{4}{c}{\texttt{Stylo-Blog}} & \multicolumn{3}{c}{\texttt{Stylo-Literary}}\\
\cmidrule(lr){2-5}\cmidrule(lr){6-9}\cmidrule(lr){10-12}
\diagbox[dir=SE,height=2\line]{Detector}{Generator} & Cohere & ChatGPT & Davinci & BLOOMZ & Qwen2.5-72B & Claude-4 & Claude-3.7 & GPT-4o & Llama3.1-8B & Phi-4 & Qwen3-4B\\
\midrule
Entropy        & 31.83 & 26.35 & 40.10 & 41.33 & 13.02 & 36.57 & 34.08 & 63.43 & 55.23 & 51.92 & 76.18\\
Lastde         & 97.69 & 97.48 & 83.70 & 88.03 & 92.69 & 68.58 & 50.96 &  6.67 & 69.88 & 65.67 & 62.57\\
Lastde++       & 98.22 & 98.67 & 84.41 & 80.91 & 99.07 & 83.37 & 88.27 & 58.41 & 60.38 & 47.57 & 39.78\\
Log-Likelihood & 93.12 & 93.76 & 72.79 & 59.75 & 95.27 & 72.30 & 77.29 & 41.65 & 36.59 & 30.94 &  9.23\\
LogRank        & 94.21 & 94.94 & 73.18 & 69.42 & 95.61 & 71.31 & 75.37 & 34.89 & 38.53 & 32.60 & 10.44\\
Detect-LRR     & 94.88 & 96.13 & 74.03 & 84.34 & 94.85 & 67.16 & 66.99 & 20.26 & 45.73 & 40.62 & 19.43\\
Fast-DetectGPT & 98.78 & 98.99 & 85.28 & 55.01 & 99.47 & 84.60 & 89.43 & 57.57 & 33.22 & 18.47 &  8.71\\
\midrule
\textbf{Avg.}
  & \cellcolor[HTML]{24a7ff} 86.96 
  & \cellcolor[HTML]{25a8ff} 86.62 
  & \cellcolor[HTML]{4cbdfa} 73.36 
  & \cellcolor[HTML]{5ac5f8} 68.40 
  & \cellcolor[HTML]{2cabfe} 84.28 
  & \cellcolor[HTML]{58c4f8} 69.13 
  & \cellcolor[HTML]{59c4f8} 68.91 
  & \cellcolor[HTML]{abf2ee} 40.41 
  & \cellcolor[HTML]{94e5f1} 48.51 
  & \cellcolor[HTML]{a9f1ee} 41.11 
  & \cellcolor[HTML]{c3ffeb} 32.33 \\
\bottomrule
\end{tabular}
}
\caption{Performance of MGT detectors, grouped by generator: shown is AUROC given one generator. 
\colorbox[HTML]{24A7FF}{Blue}: higher AUROC; \colorbox[HTML]{C3FFEB}{green}: lower AUROC.
}
\label{tab:avg-res}
\end{table*}

\subsection{Evaluation Setup}
\label{exp}
\paragraph{Evaluation Datasets} 
Apart from  \dataset for personalized scenarios, we also evaluate on an English subset of \texttt{M4}~\cite{wang-etal-2024-m4}, to show the MGT detectors' performance in a general setup.
The English subset of \texttt{M4} spans diverse sources, and contains MGTs from four generators: ChatGPT, Cohere, text-davinci-003 (Davinci), and BLOOMz, with \(\sim\)3,000 MGTs per source–generator pair. We further explain the data source and give more statistics in Appendix~\ref{app:m4}.


\paragraph{\textcolor{blue}{Baselines \& Evaluation}}
\textcolor{blue}{
We evaluate seven representative training-free detectors, which achieve strong MGT detection performances in general domain:  
Log-Likelihood~\cite{solaiman2019releasestrategiessocialimpacts},  
LogRank~\cite{solaiman2019releasestrategiessocialimpacts},  
DetectLRR~\cite{su-etal-2023-detectllm},  
Entropy~\cite{gehrmann-etal-2019-gltr,IppolitoDCE20},  
Fast-DetectGPT~\cite{bao2024fastdetectgpt},  
Lastde and Lastde++~\cite{xu2025trainingfree}.  
Details are provided in Appendix~\ref{app:baseline}.  
We use AUROC  as the evaluation metric following prior work~\cite{xu2025trainingfree,bao2024fastdetectgpt}.
}


\textcolor{blue}{
We mainly focus on training-free MGT detectors because they rely on a small set of explicit text features, making performance changes easier to interpret as feature shifts across domains and across different data distributions and evaluation conditions in practice.
Training-based detectors learn more complex representations influenced by data and model factors, so their behavior is harder to attribute to specific textual properties.
However, we also include experiments for training-based methods in Appendix~\ref{app:exp-train}.
We use AUROC to measure the overall performance of an MGT detector, as detailed in Appendix~\ref{app:auroc}.
}

\subsection{Main Results}

Table~\ref{tab:avg-res} presents the performance of various detectors across datasets. We report the average AUROC for each generator across all subdomains, with full experimental results detailed in Appendix~\ref{app:full-res}. The last row shows the average AUROC for each generator across all baselines.
The results reveal four primary findings:

(1) \textit{Significant performance degradation} occurs in personalized settings. Average AUROC on \texttt{M4} (above 85\%) falls sharply on stylized datasets, dropping to as low as 32.33\% on \texttt{Stylo-Literary}—worse than random guessing.

(2) \textit{High variance under domain shift}: Detectors exhibit divergent trends; for instance, while Entropy improves from 31.83\% in \texttt{M4} to 76.18\% in \texttt{Stylo-Literary}, Lastde drops from 97.69\% to 62.57\%.

(3) \textit{Systematic and abnormal inversions}: Many detectors experience dramatic flips, with AUROC for methods like Fast-DetectGPT falling as low as 8.71\%, indicating a near-complete reversal of discriminative capability.

(4) \textit{Increased instability in complex styles}: Fluctuations are more pronounced on \texttt{Stylo-Literary} than on \texttt{Stylo-Blog}, with abnormal reversals occurring more frequently.

These observations suggest that existing MGT detectors may be highly unstable in personalized scenarios. When generators effectively imitate specific styles, detector performance tends to shift unpredictably or even reverse entirely. Detailed per-domain performance analyses are provided in Appendix~\ref{app:perf-change}.


\section{The Feature-Inversion Trap}
Building on the observed instability in personalized scenarios, we next analyze its mechanism.
In \S\ref{inverse_trap_prem}, we introduce the feature-inversion trap hypothesis; in \S\ref{verify}, we extract the most salient inverted feature vector to verify this hypothesis; and in \S\ref{trap_general}, we demonstrate its generality across datasets.

\subsection{Feature-Inversion Trap Hypothesis}
\label{inverse_trap_prem}
\paragraph{Probing Method}
To analyze the differences between HWT and MGT, we require a representation space that effectively captures semantic and stylistic features. 
Prior work has shown that the hidden space of pretrained language models often encodes diverse linguistic and stylistic properties along approximately linear directions~\cite{mikolov2013linguistic}.
Based on this insight, we adopt GPT-2~\cite{NEURIPS2020_1457c0d6} as a proxy model, using activations from its different modules as feature representations. 
However, it remains unclear which specific modules within such a model most prominently encode task-relevant features.
We examine all modules of GPT-2, including attention layers, MLP layers, and residual streams. 

\begin{figure}[!t]
\centering
\includegraphics[width=\columnwidth]{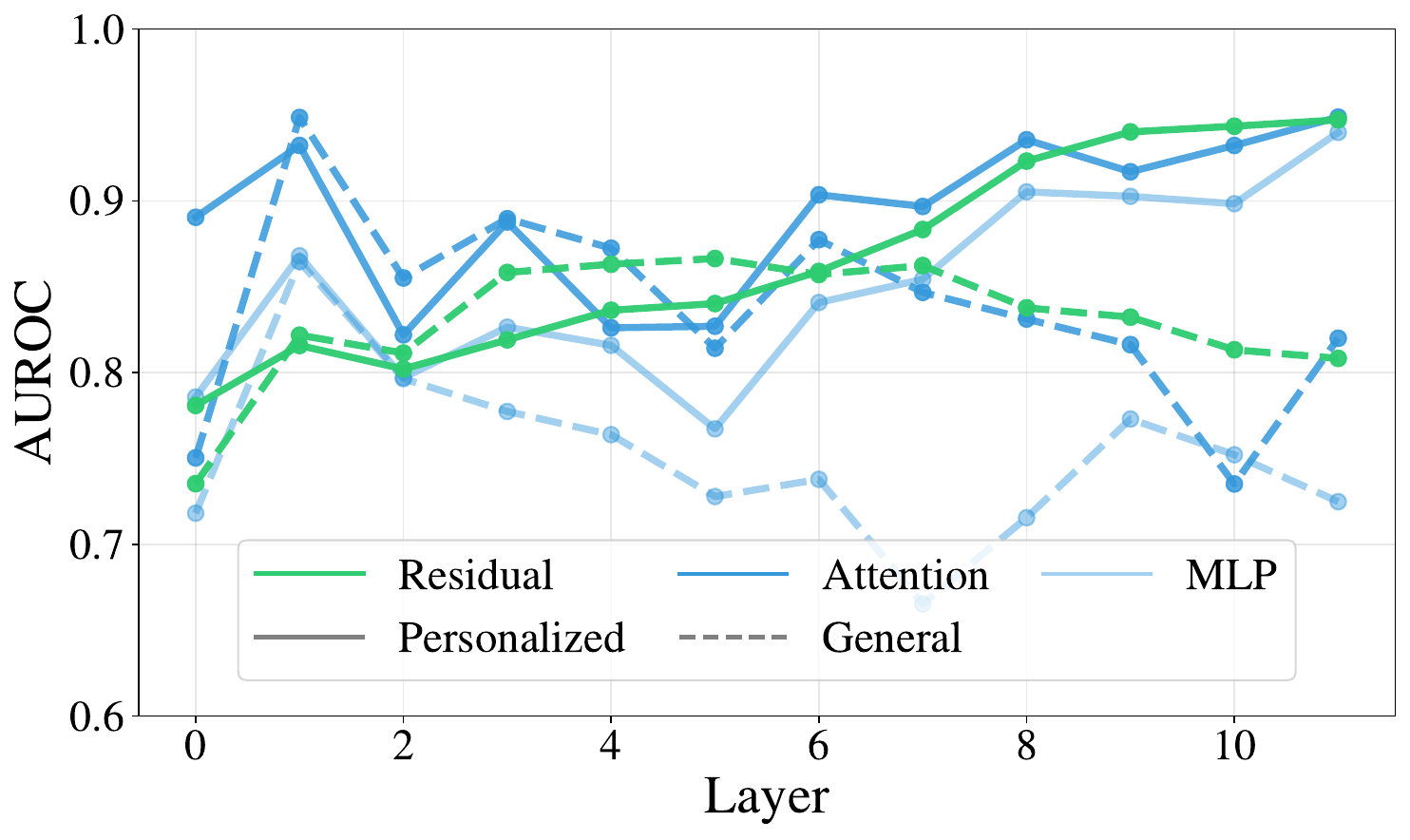} 
\caption{
AUROC scores of MGT classifiers across modules in two domains. The colors denote modules, and the line styles denote domains. We can see that the deep residual layers are best at distinguishing HWT/MGT in both domains.
}
\label{fig:raw-auc}
\end{figure}

For each module, we extract the activation of the last token as the text representation and train a logistic regression classifier to distinguish HWT from MGT in general and personalized domains.
Figure~\ref{fig:raw-auc} reports the AUROC for each module.
Deep residual streams consistently achieve high AUROC across the two domains,
indicating that they retain strong discriminative features. 
Therefore, we focus on the residual stream at a near-final layer, i.e., layer 10, in the following analysis.

\paragraph{Probe Datasets} 
We construct two small probe datasets. Following~\cite{xu2025trainingfree}, we adapt Xsum~\cite{narayan-etal-2018-dont} and their LLM-generated continuations as the general-domain data. 
We randomly sample 150 texts from the J.A subset of \texttt{Stylo-Literary} and take their Phi-4 MGTs as personalized-domain data. 
Each domain contains 150 HWT and 150 MGT samples.



\paragraph{Visibility of the Inverting Trend}

Since the detector's performance fluctuates under domain shift, we first investigate whether the representations themselves encode domain differences across varying data distributions and stylistic conditions.


Following~\cite{gao2025evaluatebiasmanualtest}, we train a logistic regression-style domain classifier, whose weight direction serves as domain-related features, to distinguish general HWT from personalized HWT, and then evaluate it on HWT/MGT samples in both domains.
Figure~\ref{fig:logit_distri}(a) shows a clear separation of the projection onto the weight direction (feature values) between the two domains, as expected.
Notably, the feature value of MGT is lower than HWT in the general domain but higher in the personalized domain, revealing a consistent cross-domain reversal pattern in feature behavior.
This leads to the \textit{feature-inversion trap hypothesis}: distinguishable MGT features in general domain are inverted under personalization.


\subsection{Verification of the Feature-Inversion Trap}
\label{verify}


In order to verify the existence of the feature-inversion trap, we aim to identify the most representative inverted feature direction that consistently captures cross-domain discrepancies in feature behavior.
If the projection of datasets in this direction is highly correlated with detector performance, it would suggest that detectors rely on it, thereby supporting our hypothesis.

\subsubsection{Deriving the Inverted Feature Direction}
\label{get_direction}
We begin by extracting the inverted feature direction that is most responsible for this effect, in order to assess its correlation with detector performance.

\label{invert_direction}
\paragraph{Notation} 
We denote the general-domain dataset by $G$ and the personalized-domain dataset by $S$. 
Let $g_+, g_- \in G$ and $s_+, s_- \in S$ represent MGT and HWT activations in the two domains. 
For each quadruple $(g_+, g_-, s_+, s_-)$, we compute domain-specific difference vectors:
\begin{align}
    v_G = g_+ - g_-, \quad v_S = s_+ - s_-.
\end{align}

\begin{figure}[!t]
\centering
\includegraphics[width=\columnwidth]{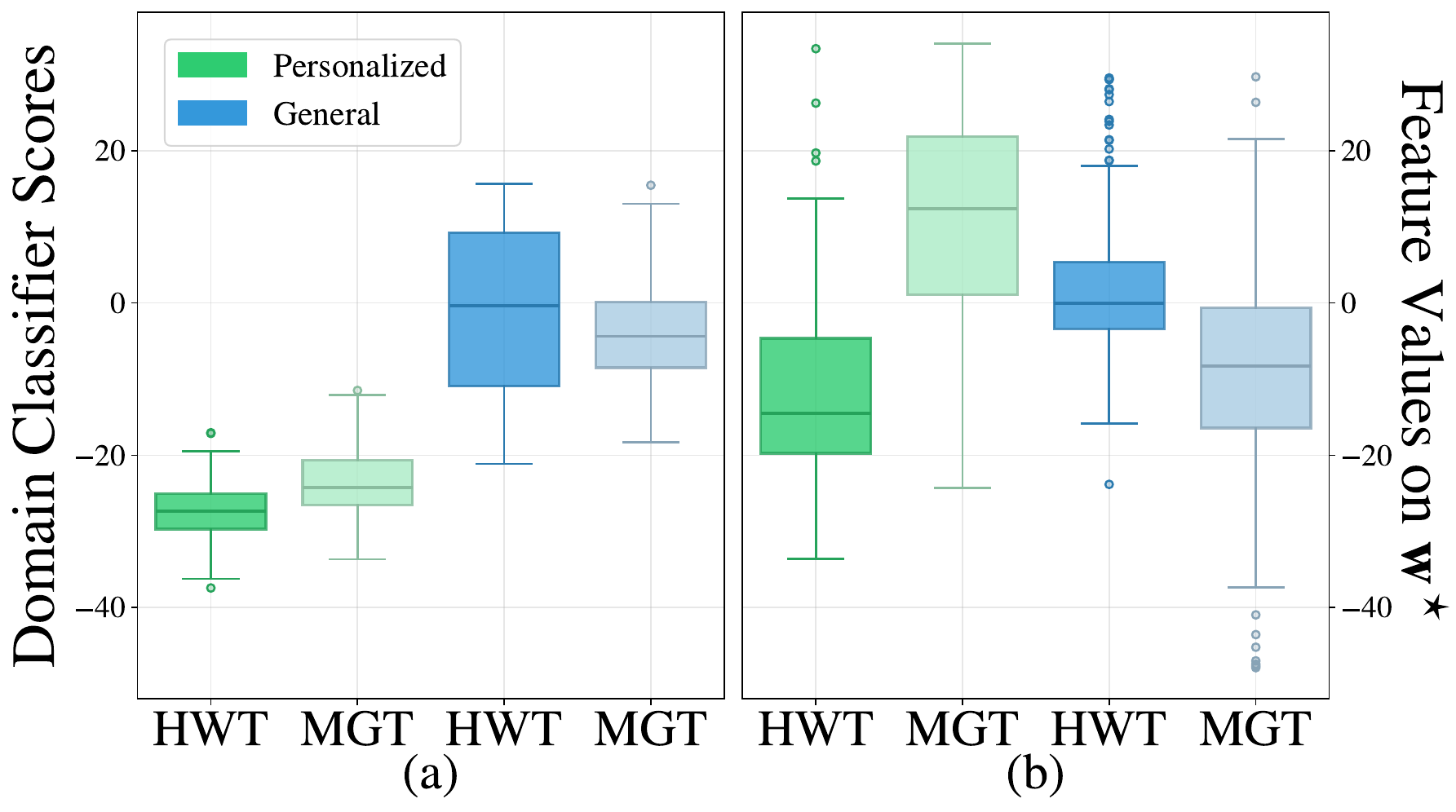} 
\caption{
(a) Domain classifier score distribution: We can see moderate inversion effects.  
(b) Feature value distributions on the inverted feature direction \(\mathbf{w}^\star\): we can see a major inversion effect.
}
\label{fig:logit_distri}
\end{figure}

\paragraph{Inversion-Value Matrix and Object}
Our goal is to find a direction $\mathbf{w}$ where the projection of $v_G$ is opposite to that of $v_S$.
For each quadruple $(g_+, g_-, s_+, s_-)$, we define the projection product in direction $\mathbf{w}$ as
\begin{equation}
\label{eqi}
q_i(\mathbf{w})=(\mathbf{w}^\top v_G)(\mathbf{w}^\top v_S)=\mathbf{w}^\top (v_G v_S^\top)\mathbf{w}.
\end{equation}
Since for any matrix \(\mathrm{M}\), it holds that $\mathbf{w}^\top \mathrm{M} \mathbf{w} = \mathbf{w}^\top \tfrac{1}{2}\mathrm{(M+M^\top)}\mathbf{w}$, Equation~\ref{eqi} can be rewritten as 
\begin{equation}
\label{get_a}
    \scalebox{1}{$q_i(\mathbf{w})
    = \textstyle \mathbf{w}^\top \tfrac{1}{2}\bigl(v_G v_S^\top + v_S v_G^\top\bigr)\mathbf{w}.$}
\end{equation}
Further details on the calculation of Equation~\ref{eqi} and~\ref{get_a} are in Appendix~\ref{app:rayleigh}.
For each quadruple, we define a cross-domain matrix
\begin{equation}
    A_i=\textstyle \frac{1}{2}\Bigl(v_G v_S^\top + v_S v_G^\top\Bigr),
\end{equation}
which is symmetry. We aggregate over quadruples:
\begin{equation}
A=\textstyle \sum_i A_i.
\label{eq:Arev}
\end{equation}
The overall inversion objective then is
\begin{equation}
\resizebox{0.4\textwidth}{!}{$
\mathcal{R}(\mathbf{w})=\textstyle\sum_i q_i(\mathbf{w})
=\mathbf{w}^\top A\mathbf{w},\quad \text{s.t.}\ ||\mathbf{w}||=1.
$}
\label{eq:Rrev}
\end{equation}
Since each $A_i$ is symmetric, the aggregated matrix $A$ in Equation~\ref{eq:Arev} is also symmetric. 
So far, we have transformed the problem into the \textit{Rayleigh quotient} of $A$ with respect to $\mathbf{w}$ under the unit-norm constraint.  Illustrations on the Rayleigh quotient problem are available in Appendix~\ref{app:rayleigh}.

\paragraph{Solution}
By the property of the Rayleigh quotient, minimizing the objective in Eq.~\ref{eq:Rrev} reduces to $\mathbf{w}^\star=\textstyle \arg\min_{|\mathbf{w}|=1}\mathcal{R}(\mathbf{w})$,
whose the solution is the eigenvector of $A$ that corresponds to its smallest eigenvalue: $A=U\Sigma U^\top,\quad\mathbf{w}^\star = U[:, -1]$,
where $U[:, -1]$ denotes the last column of $U$, associated with the minimum eigenvalue. 

This $\mathbf{w}^\star$ represents the \textit{inverted feature direction}, i.e., the axis along which the HWT–MGT projection (feature value) difference in the general domain is most strongly inverted in the personalized domain.
Using the probe datasets in~\S\ref{inverse_trap_prem}, we show feature value distributions of four sample types on $\mathbf{w}^\star$ in Figure~\ref{fig:logit_distri}(b).
In the personalized domain, MGT values are lower than HWT, while in the general domain, MGT values are higher.
This flip provides direct evidence of feature inversion.

\begin{figure}[!t]
\centering
\includegraphics[width=\columnwidth]{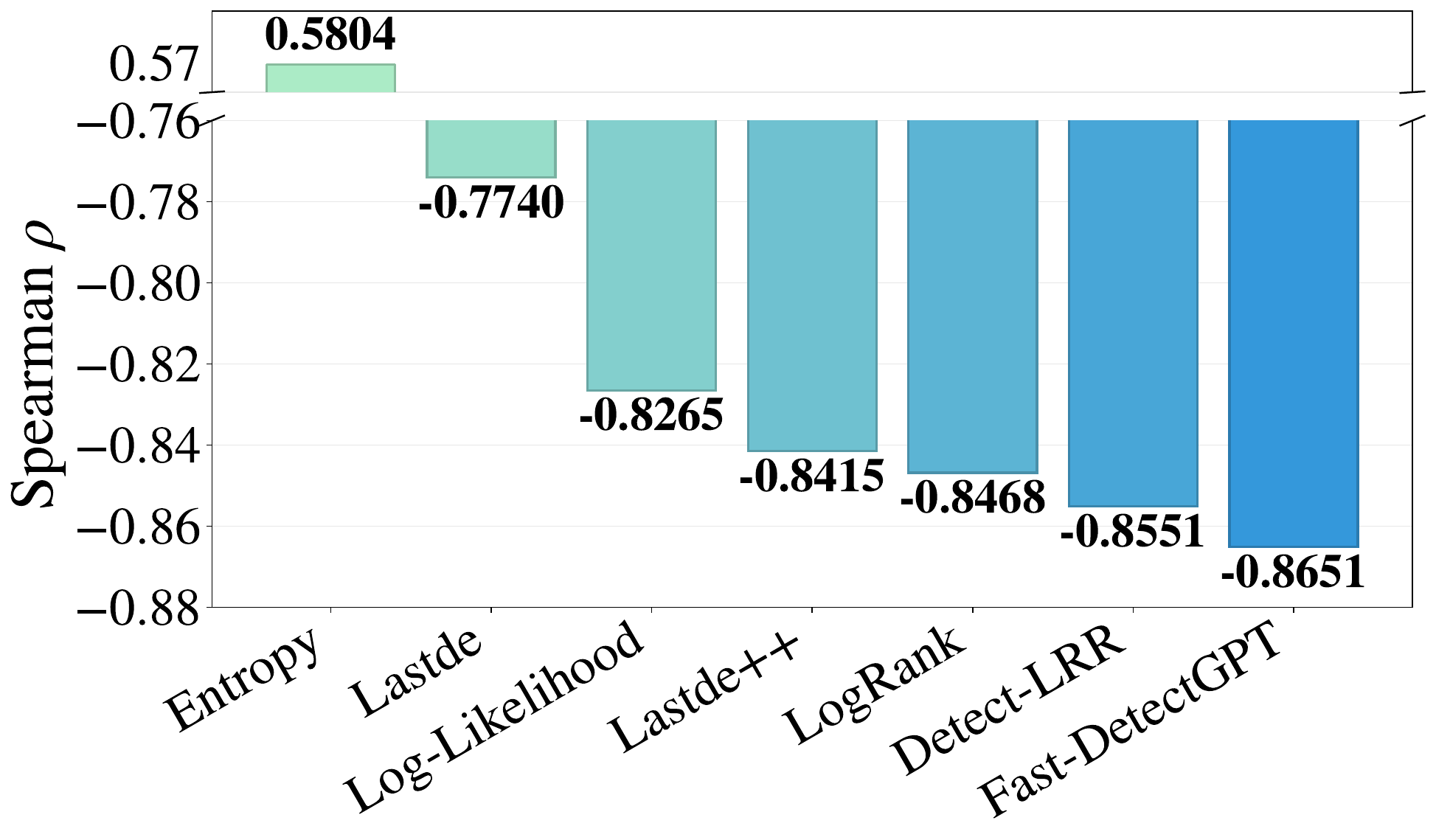} 
\caption{
Spearman \(\rho\) between feature value differences of datasets and corresponding detector performance.
}
\label{fig:corr-distri}
\end{figure}




\subsubsection{Correlation with Detector Performance}
\label{corr}
Deriving the inverted feature direction $\mathbf{w}^\star$ reveals a dimension where MGT and HWT roles flip across domains, but this alone does not confirm its effect on detectors. To establish the connection, we evaluate the correlation between the strength of the inverted feature and detector performance.

Intuitively, along \(\mathbf{w}^\star\), the relative positions of HWT and MGT feature values in the two domains are inverted. We quantify this property using \textit{feature value difference}.
For a dataset \(M\) with its MGT denoted as \(m_+\) and HWT as \(m_-\), the feature value difference is
\begin{equation}
\scalebox{0.85}{$
   \mathcal{D}(M,\mathbf{w}^\star)=\sum_{\{m_+,m_-\}\subset M}\Bigl(m_+^\top\mathbf{w}^{\star}-m_-^\top\mathbf{w}^{\star}\Bigr),
$}
\end{equation}
which reflects the overall discrepancy between MGTs and HWTs on the inverted feature. Larger values indicate a clearer separation, while smaller or flipped values suggest confusion between the two classes.
Following the experimental design in~\S\ref{exp}, we partition \texttt{M4} and \dataset by unique generator–subfield combinations, resulting in a total of \(N=45\) subsets.
 For each subset, we compute \(\mathcal{D}(\cdot,\mathbf{w}^\star)\), forming a set \(\{\mathcal{D}_i\}_{i=1}^N\). Meanwhile, for each MGT detector, we collect the AUROCs on the same subsets, denoted as \(\{\mathrm{AUROC}_i\}_{i=1}^N\). We measure their consistency by Spearman correlation: $\rho = \text{Spearman}\big(\{\mathcal{D}_i\}, \{\mathrm{AUROC}_i\}\big)$.

The resulting correlations \(\rho\) for each detector are shown in Figure~\ref{fig:corr-distri}.
We can see that entropy exhibits a positive correlation (\(\sim\) 0.6), whereas all other detectors have  \(\rho<-0.77\), indicating strong negative correlations. 
We further show the distribution of \(\{\mathrm{AUROC}_i\}\) versus \(\{\mathcal{D}_i\}\) of each  detector in Appendix~\ref{app:corr-viz}.
Overall, these results demonstrate that the detector's performance is tightly linked to a feature inverted across domains. 

\textcolor{blue}{
To verify that this correlation is not spurious, we further conduct experiments that isolate the effects of inverted features. We evaluate detectors on randomized text lacking semantic content, where positive and negative samples are separated along the inverted direction, an orthogonal direction, or at random.  
Detectors only show strong discrimination under the first case, indicating their direct reliance on inverted features (Appendix~\ref{app:dep_verif}).
}

\label{trap_general}
\begin{figure}[!t]
\centering
\includegraphics[width=\columnwidth]{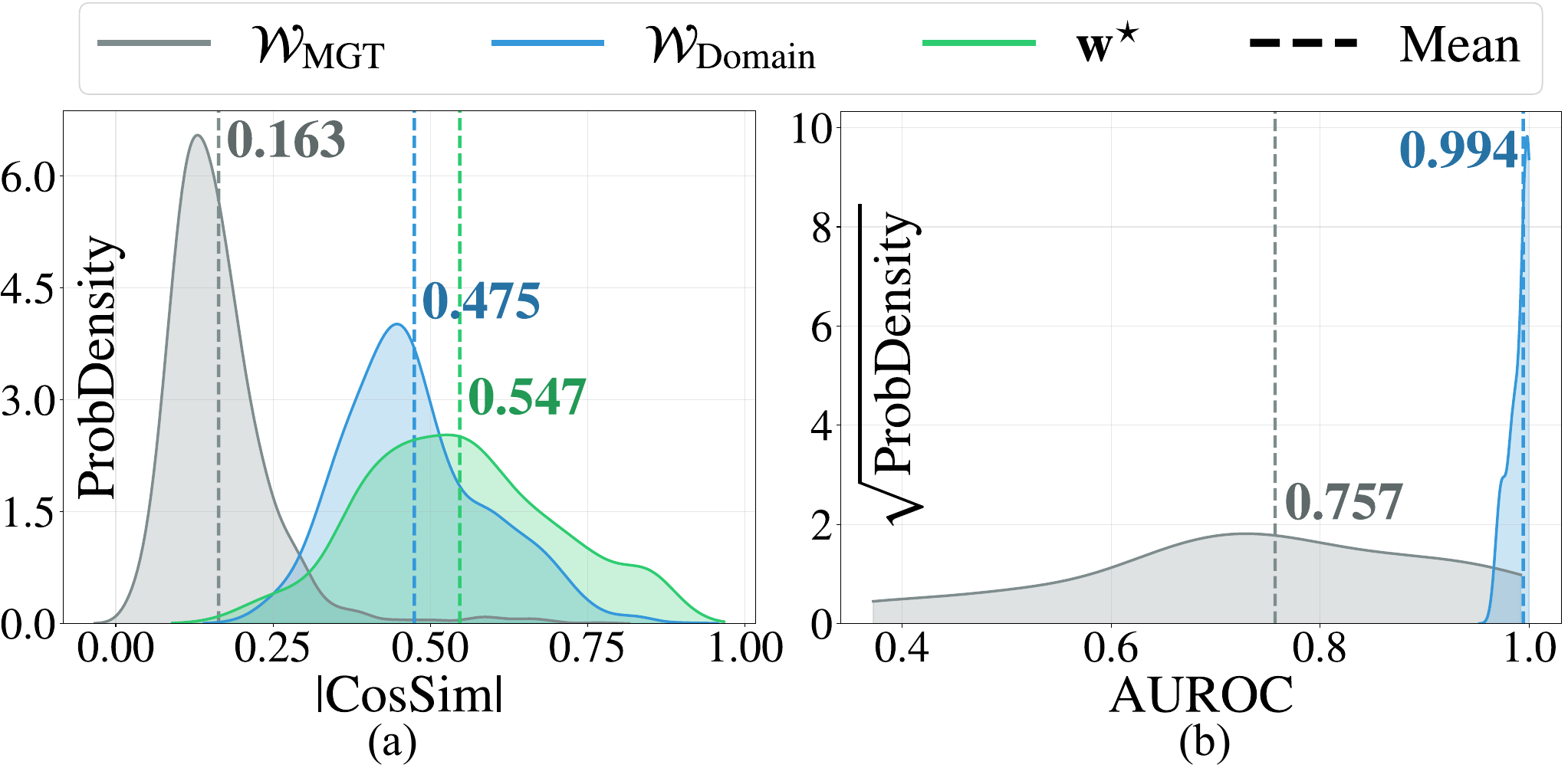} 
\caption{
(a) \(|\mathrm{CosSim}|\) between feature directions from different datasets. \(\mathbf{w}^\star\) are close to \(\mathcal{W}_{\mathrm{Domain}}\) and beyond \(\mathcal{W}_\mathrm{MGT}\). 
(b) AUROC distributions of the generalization test. \(\mathcal{W}_{\mathrm{Domain}}\) has evidently better generalizability.
}
\label{fig:cossims}
\end{figure}

\subsection{Generality of the Feature-Inversion Trap}

Having verified that \(\mathbf{w}^\star\) captures a key inverted feature correlated with performance, we now investigate whether this phenomenon is dataset-specific or reflects a broader, cross-domain pattern.
To this end, we evaluate the consistency of inverted features across multiple datasets.
We processed \texttt{Stylo-Literary} and \texttt{M4} as follows. For each subdomain–generator pair in \texttt{M4}, we create five subsets, each with 150 random HWTs and 150 MGTs, as general-domain probe datasets. 


Also, we apply the same procedure to \texttt{Stylo-Literary} to obtain personalized-domain probe datasets for consistent comparison and reliable cross-domain evaluation across sampling trials.
Each experiment samples one general subset and one personalized subset. We then extract three feature directions: (1) the inverted feature direction $\mathbf{w}^\star$, (2) the MGT feature direction $\mathcal{W}_{\mathrm{MGT}}$, and (3) the domain feature direction $\mathcal{W}_{\mathrm{Domain}}$.
 
To obtain these two reference directions, we train logistic regression models. One model separates HWT and MGT and gives $\mathcal{W}_{\mathrm{MGT}}$. The other separates general and personalized data and gives $\mathcal{W}_{\mathrm{Domain}}$. We repeat this process 100 times and produce 100 sets of the three types of vectors. We then compute cosine similarity within each group.

Figure~\ref{fig:cossims}(a) shows that $\mathcal{W}_{\mathrm{MGT}}$ has low similarity with a mean of 0.163. $\mathcal{W}_{\mathrm{Domain}}$ and $\mathbf{w}^\star$ show higher stability with means of 0.475 and 0.547.
We also test how $\mathcal{W}_{\mathrm{MGT}}$ and $\mathcal{W}_{\mathrm{Domain}}$ generalize to new and previously unseen subsets. 

Figure~\ref{fig:cossims}(b) shows that $\mathcal{W}_{\mathrm{Domain}}$ keeps a high AUROC in other related subsets with a mean of 0.994, while $\mathcal{W}_{\mathrm{MGT}}$ varies more widely from 0.4 to 0.8 with a mean value of 0.757. The strong overall similarity between $\mathcal{W}_{\mathrm{Domain}}$ and $\mathbf{w}^\star$ indicates that inverted features share the same high level of generalization ability.

\textcolor{blue}{          
Based on these observations, we conclude that the feature-inversion trap is a widespread phenomenon between personalized and general domains, and the inverted features share strong commonalities across various different datasets. However, general MGT detectors often suffer from the feature-inversion trap, leading to unreliable performance in personalized domains. So how can we effectively leverage such consistent inversion to enhance detector transferability?
}

\section{\method}

In this section, we propose \method, an automatic transferability estimator that predicts performance shifts by quantifying detectors’ reliance on inverted features. 
Instead of directly measuring transfer performance, it evaluates detector behavior on controlled probe datasets that isolate these features, providing an indirect yet effective estimate of transferability.

\begin{figure}[!t]
\centering
\includegraphics[width=\columnwidth]{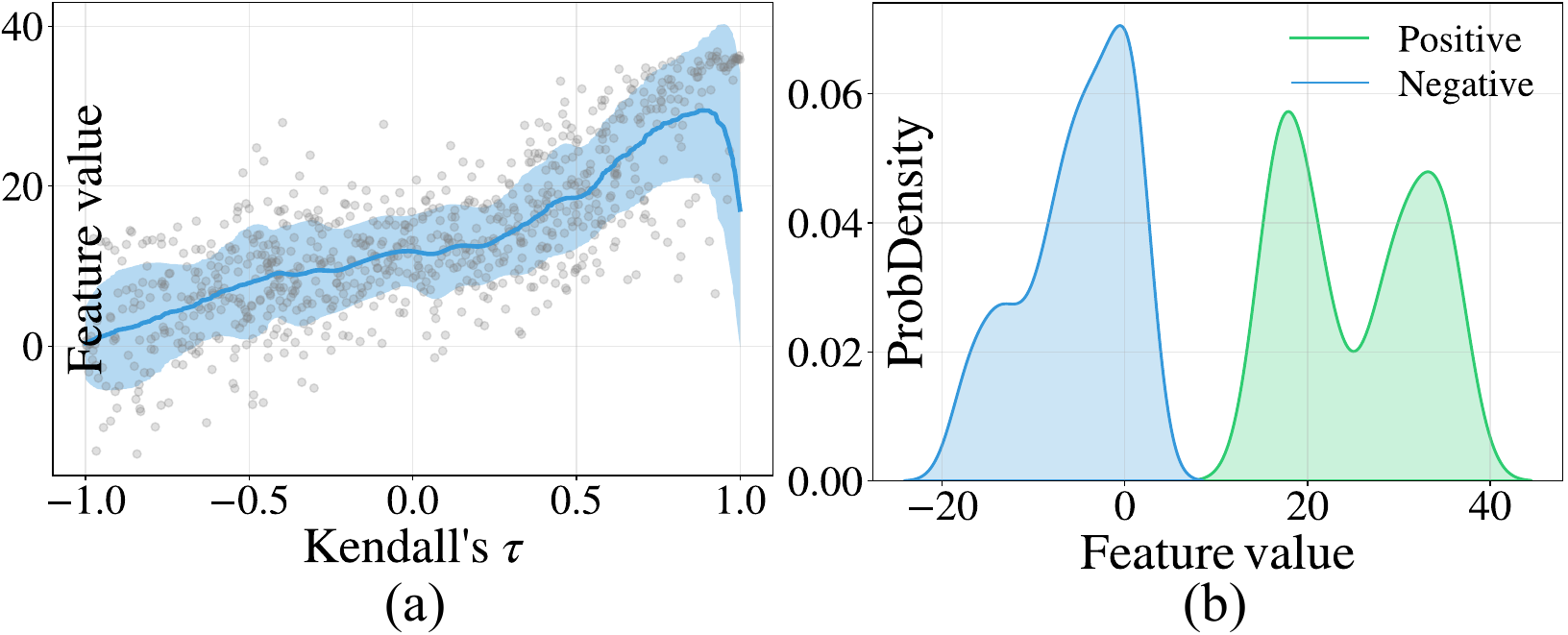} 
\caption{
(a) Changes of feature values under different levels of token shuffling. Shuffling brings consistent change in feature value. (b) Distribution of feature values in the probe dataset. Two classes share zero overlap.
}
\label{fig:scramble}
\end{figure}

\subsection{Design of \method}
\subsubsection{Probe Dataset Synthesis}  
To construct probe datasets that differ only in inverted features, we eliminate confounding factors from text semantics, domain, and class (HWT/MGT) by shuffling tokens.
To control shuffle strength, we use Kendall's $\tau$, a measure of sequence order ranging from 1 to $-1$, corresponding to a gradual inversion of token order (see Appendix~\ref{app:corr-coef}).
For each sentence, we generate variants with Kendall’s $\tau$ spanning this range.
As shown in Figure~\ref{fig:scramble}(a), both Kendall's $\tau$ and the corresponding feature values vary continuously.

\textcolor{blue}{Therefore, we build probe datasets by shuffling tokens.
We sample one general and one personalized HWT, generate 800 variants for each with different Kendall's $\tau$, merge them, and select the 50 samples with the highest feature values as positives and the 50 lowest as negatives.
As shown in Figure~\ref{fig:scramble}(b), the resulting feature value distributions show no overlap.
We further evaluate the style and MGT linear classifiers introduced in~\S\ref{inverse_trap_prem}, which achieve near-perfect accuracy during training but drop to 53\% and 66\% AUROC on the probe dataset, respectively, confirming the effective removal of domain and class features.
The probe dataset thus reflects only differences in the inverted features.
}


\subsubsection{Transferability Evaluation}  

\begin{figure}[!t]
\centering
\includegraphics[width=\columnwidth]{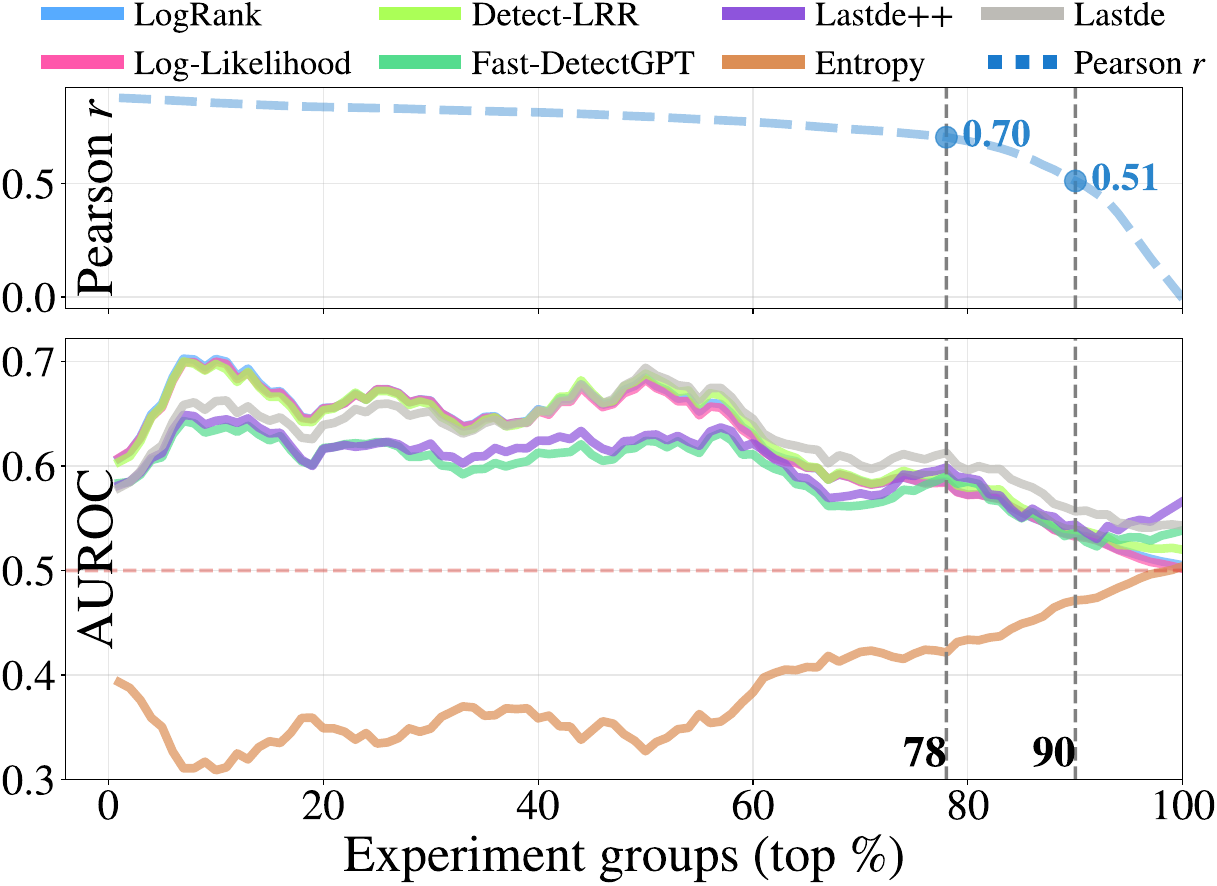} 
\caption{
Top: Pearson \(r\) between transfer gaps and AUROCs.
Bottom: corresponding AUROCs of detectors in probe datasets.
 Percentages of experiments groups where \(r>0.7\) and \(r>0.5\) are marked.  
}
\label{fig:scramble_corr}
\end{figure}

We next describe how detector performance on the probe dataset reflects transferability. 
Since the probe dataset removes semantic, domain, and class-related signals, performance on it more directly and clearly reveals the role of inverted features in detector decisions.

Our evaluated detectors treat MGT as positive samples, which in both the general domain and the probe dataset tend to have higher feature values.
Therefore, if a detector relies on the inverted feature, it should perform well on the probe dataset. 
AUROC reflects the degree of reliance on inverted features: high values indicate strong dependence and degradation after transfer, low values suggest inverted dependence and potential performance gains, and values near 0.5 imply weak dependence and stable transfer.

\subsection{Performance of \method}
We test seven detectors to examine how well it reflects their performance shifts during transfer.

\paragraph{Evaluation Setup.}  
We construct 100 probe datasets from \texttt{M4} and \texttt{Stylo-Literary}. In each experiment, we randomly choose five of them for testing, and we compute the mean AUROC. We then measure the Pearson \(r\) between this mean AUROC and the detector's overall performance gap between \texttt{M4} and \texttt{Stylo-Literary}. The results of over 100 such experiments are shown in Figure~\ref{fig:scramble_corr}.

\paragraph{Results.}  
\label{main_res}

Figure~\ref{fig:scramble_corr} shows the AUROC of all detectors across experiments. Two patterns appear. 
(1) Entropy stays below 0.5 in all runs, showing inverted reliance, while all other detectors remain above 0.5, showing positive reliance. This matches their transfer behavior, where Entropy improves, and others degrade. 
(2) In 90\% of runs, Pearson \(r\) exceeds 0.5, and in 78\%\ it exceeds 0.7, indicating stable reliance levels. 
These results show that \method identifies detectors' dependence on inverted features and also captures how strong that dependence is. 
\textcolor{blue}{We add an ablation study on the number of probe datasets in Appendix~\ref{app:abl}.}




\section{\textcolor{blue}{Discussion}}
\label{disc}
\textcolor{blue}{In this section, we address several conceptual questions raised by the observed mechanism, with empirical evidence deferred to Appendix~\ref{app:evd}.}

\paragraph{\textcolor{blue}{(i) Is the feature-inversion trap a typical out-of-distribution (OOD) effect?}}
\textcolor{blue}{The feature-inversion trap is a special case of OOD, marked by two points: It aligns with the inverted feature direction, and it often causes a reversal of detector behavior rather than simple degradation. Appendix~\ref{app:ood-exp} shows that common OOD  do not produce these patterns. Appendix~\ref{app:con-terms} also outlines its links to related terms such as spurious correlations.}

\paragraph{\textcolor{blue}{(ii) What do inverted features capture?}}
\textcolor{blue}{
Our evidence shows that they relate to text diversity. Many training-free detectors~\cite{xu2025trainingfree,gehrmann-etal-2019-gltr} assume that HWT is more diverse than MGT, but personalization breaks this pattern: personalized MGT can be more varied and less coherent. This shift is consistent with the latent we observe and helps explain the negative performance flips of these detectors. See Appendix~\ref{app:what-inv-feat}.}

\paragraph{\textcolor{blue}{(iii) How can we mitigate the feature-inversion trap?}}
\textcolor{blue}{
A practical option is to use tuned training-based detectors. They learn more cues and can reach strong in-domain accuracy after training on personalized text, though their cross-domain generalization remains limited, as shown in Appendix~\ref{app:mitigate}. For training-free detectors, using features less sensitive to style drift, such as stable traits of human writing, may reduce reliance on diversity signals. Adaptive thresholding~\cite{jung2025groupadaptivethresholdoptimizationrobust} that adjusts to the stylization can also improve robustness in personalized settings.
}

\section{Conclusion and Future Work}
\textcolor{blue}{We presented \dataset, the first dataset for MGT detection in personalized scenarios. Our study showed that existing detectors face large performance shifts, and even inversion, after domain transfer. We traced this to the feature-inversion trap, where features that separate MGT and HWT change their roles across domains and lead detectors to flip predictions. Based on this, we proposed \method, a transferability framework that measures how much detectors rely on inverted features. 

In future work, we plan to explore MGT detection methods that avoid such features to support stronger transferability.}


\section*{Limitations}
Our study focuses on English, and further work is needed to assess whether the findings generalize to multilingual, domain-specific, or code-switched settings. Linguistic variation may introduce different stylistic cues and distributional properties that affect both personalization and detection behavior beyond the current analysis.

While feature inversion explains much of the performance shift under personalization, other latent factors may also affect robustness. Aspects such as discourse structure or higher-level semantics are not modeled and remain to be explored.

Finally, the experiments are conducted in controlled offline settings with static benchmarks, which may not fully capture real-world scenarios. In practice, detection often occurs in dynamic and interactive contexts, such as adaptive generation, human–AI coauthoring, or adversarial style imitation. Evaluating under these conditions may provide a more complete understanding of robustness.

\section*{Ethical Considerations}
This work aims to advance understanding of machine-generated text (MGT) detection in personalized scenarios and is intended for research on transparency, robustness, and responsible AI use. All datasets used in our experiments are derived from publicly available sources, and no private, sensitive, or personally identifiable information is included. The generation process follows open and reproducible settings without targeting any real individuals. While our findings reveal potential weaknesses in existing detectors, they are presented to support the development of safer and more reliable detection systems rather than to facilitate misuse or impersonation. We encourage future research to apply these insights ethically, ensuring that detection technologies are used to mitigate misinformation and protect authorship integrity.

\bibliography{custom}

@article{abdin2024phi4technicalreport,
 archivePrefix = {arXiv},
 author = {Marah Abdin and Jyoti Aneja and Harkirat Behl and Sébastien Bubeck and Ronen Eldan and Suriya Gunasekar and Michael Harrison and Russell J. Hewett and Mojan Javaheripi and Piero Kauffmann and James R. Lee and Yin Tat Lee and Yuanzhi Li and Weishung Liu and Caio C. T. Mendes and Anh Nguyen and Eric Price and Gustavo de Rosa and Olli Saarikivi and Adil Salim and Shital Shah and Xin Wang and Rachel Ward and Yue Wu and Dingli Yu and Cyril Zhang and Yi Zhang},
 journal = {ArXiv preprint},
 volume = {arXiv:2412.08905},
 eprint = {2412.08905},
 primaryClass = {cs.CL},
 title = {Phi-4 Technical Report},
 year = {2024}
}

@article{abs-2402-12715,
  author       = {Wenqian Ye and
                  Guangtao Zheng and
                  Xu Cao and
                  Yunsheng Ma and
                  Xia Hu and
                  Aidong Zhang},
  title        = {Spurious Correlations in Machine Learning: {A} Survey},
  journal = {ArXiv preprint},
  volume = {arXiv:2402.12715},
  eprint = {2402.12715},
  year         = {2024},
  archivePrefix = {arXiv}
}

@article{ZhouLQXL23,
 author = {Kaiyang Zhou and
Ziwei Liu and
Yu Qiao and
Tao Xiang and
Chen Change Loy},
 title = {Domain Generalization: {A} Survey},
 journal = {{IEEE} Transactions on Pattern Analysis and Machine Intelligence},
 volume={45},
 number={4},
 pages={4396--4415},
 year={2022},
 publisher={IEEE}
}

@inproceedings{IlyasSTETM19,
 author = {Andrew Ilyas and
Shibani Santurkar and
Dimitris Tsipras and
Logan Engstrom and
Brandon Tran and
Aleksander Madry},
 booktitle = {Proceedings of the Conference on Neural Information Processing Systems},
 series = {NeurIPS~'19},
 address = {Vancouver, Canada},
 pages = {125--136},
 title = {Adversarial Examples Are Not Bugs, They Are Features},
 year = {2019}
}

@inproceedings{WangMISSTAM0AAH24,
 author = {Yuxia Wang and
Jonibek Mansurov and
Petar Ivanov and
Jinyan Su and
Artem Shelmanov and
Akim Tsvigun and
Osama Mohammed Afzal and
Tarek Mahmoud and
Giovanni Puccetti and
Thomas Arnold and
Alham Fikri Aji and
Nizar Habash and
Iryna Gurevych and
Preslav Nakov},
 booktitle = {Proceedings of the 62nd Annual Meeting of the Association for Computational Linguistics (Volume 1: Long Papers)},
 series = {ACL~'24},
 NOmonth = aug,
 address = {Bangkok, Thailand},
 pages = {3964--3992},
 title = {{M}4{GT}-Bench: Evaluation Benchmark for Black-Box Machine-Generated TextDetection},
 doi = "10.18653/v1/2024.acl-long.218",
 year = {2024}
}

@article{irl,
 author = {Han Zhao and
Chen Dan and
Bryon Aragam and
Tommi S. Jaakkola and
Geoffrey J. Gordon and
Pradeep Ravikumar},
 journal = {Journal of Machine Learning Research},
 volume={23},
 number={340},
 pages={1--49},
 title = {Fundamental Limits and Tradeoffs in Invariant Representation Learning},
 year = {2022}
}

@inproceedings{morris-etal-2023-text,
 author = {Morris, John  and
Kuleshov, Volodymyr  and
Shmatikov, Vitaly  and
Rush, Alexander},
 booktitle = {Proceedings of the 2023 Conference on Empirical Methods in Natural Language Processing},
 series = {EMNLP~'23},
 NOmonth = dec,
 address = {Singapore},
 doi = "10.18653/v1/2023.emnlp-main.765",
 pages = {12448--12460},
 title = {Text Embeddings Reveal (Almost) As Much As Text},
 year = {2023}
}

@inproceedings{bao2024fastdetectgpt,
 author = {Guangsheng Bao and Yanbin Zhao and Zhiyang Teng and Linyi Yang and Yue Zhang},
 booktitle = {Proceedings of International Conference on Learning Representations},
 series = {ICLR~'24},
 address = {Vienna, Austria},
 title = {{F}ast-{D}etect{GPT}: Efficient Zero-Shot Detection of Machine-Generated Text via Conditional Probability Curvature},
 year = {2024}
}

@inproceedings{dong2024rayleigh,
 author = {Xiangyu Dong and Xingyi Zhang and Sibo Wang},
 booktitle = {Proceedings of the International Conference on Learning Representations},
 series    = {ICLR~'24},
 address   = {Vienna, Austria},
 title = {Rayleigh Quotient Graph Neural Networks for Graph-level Anomaly Detection},
 year = {2024}
}

@inproceedings{DuganHTZLXIC24,
 author = {Liam Dugan and
Alyssa Hwang and
Filip Trhl{\'{\i}}k and
Andrew Zhu and
Josh Magnus Ludan and
Hainiu Xu and
Daphne Ippolito and
Chris Callison{-}Burch},
 booktitle = {Proceedings of the 62nd Annual Meeting of the Association for Computational Linguistics (Volume 1: Long Papers)},
  series    = {ACL~'24},
  NOmonth = aug,
  address   = {Bangkok, Thailand},
 title = {{RAID:} {A} Shared Benchmark for Robust Evaluation of Machine-Generated
Text Detectors},
 doi = "10.18653/v1/2024.acl-long.674",
 pages = "12463--12492",
 year = {2024}
}

@article{e22010126,
 author = {Gerlach, Martin and Font-Clos, Francesc},
 journal = {Entropy},
 number = {1},
 pages = {126},
 title = {A Standardized Project Gutenberg Corpus for Statistical Analysis of Natural Language and Quantitative Linguistics},
 volume = {22},
 year = {2020}
}

@article{gao2025evaluatebiasmanualtest,
 archivePrefix = {arXiv},
 author = {Lang Gao and Kaiyang Wan and Wei Liu and Chenxi Wang and Zirui Song and Zixiang Xu and Yanbo Wang and Veselin Stoyanov and Xiuying Chen},
 journal = {ArXiv preprint}, 
 volume = {arXiv:2505.15524},
 eprint = {2505.15524},
 primaryClass = {cs.CL},
 title = {Evaluate Bias without Manual Test Sets: A Concept Representation Perspective for LLMs},
 year = {2025}
}

@inproceedings{gehrmann-etal-2019-gltr,
 author = {Gehrmann, Sebastian  and
Strobelt, Hendrik  and
Rush, Alexander},
 booktitle = {Proceedings of the 57th Annual Meeting of the Association for Computational Linguistics: System Demonstrations},
 series = {ACL~'19},
 NOmonth = jul,
 address = {Florence, Italy},
 title = {{GLTR}: Statistical Detection and Visualization of Generated Text},
 year = {2019},
 doi = "10.18653/v1/P19-3019",
 pages = "111--116"
}

@misc{gpt-j,
 author = {Wang, Ben and Komatsuzaki, Aran},
 title = {{GPT-J-6B: A 6 Billion Parameter Autoregressive Language Model}},
 year = {2021}
}

@article{guo-etal-2023-hc3,
 author = {Guo, Biyang  and
Zhang, Xin  and
Wang, Ziyuan  and
Jiang, Minqi  and
Nie, Jinran  and
Ding, Yuxuan  and
Yue, Jianwei  and
Wu, Yupeng},
 journal = {ArXiv preprint},
 volume = {arXiv:2301.07597},
 eprint = {2301.07597},
 title = {{How Close is {ChatGPT} to Human Experts? Comparison Corpus, Evaluation, and Detection}},
 primaryClass = {cs.CL},
 year = {2023},
 archivePrefix = {arXiv}
}

@inproceedings{guo2024detective,
 author = {Guo, Xun and He, Yongxin and Zhang, Shan and Zhang, Ting and Feng, Wanquan and Huang, Haibin and Ma, Chongyang},
 booktitle = {Proceedings of the Conference on Neural Information Processing Systems},
 series = {NeurIPS~'24},
 address = {Vancouver, Canada},
 pages = {88320--88347},
 title = {{{DeTeCtive}: Detecting {AI}-generated Text via Multi-level Contrastive Learning}},
 year = {2024}
}

@inproceedings{hu2022lora,
 author = {Edward J Hu and yelong shen and Phillip Wallis and Zeyuan Allen-Zhu and Yuanzhi Li and Shean Wang and Lu Wang and Weizhu Chen},
 booktitle = {Proceedings of the International Conference on Learning Representations},
 series = {ICLR~'22},
 title = {Lo{RA}: Low-Rank Adaptation of Large Language Models},
 year = {2022}
}

@article{huang2025trustworthiness,
 author = {Huang, Yue and Gao, Chujie and Wu, Siyuan and Wang, Haoran and Wang, Xiangqi and Zhou, Yujun and Wang, Yanbo and Ye, Jiayi and Shi, Jiawen and Zhang, Qihui and others},
 journal = {ArXiv preprint}, 
 volume = {arXiv:2502.14296},
 eprint = {2502.14296},
 primaryClass = {cs.CY},
 title = {On the trustworthiness of generative foundation models: Guideline, assessment, and perspective},
 year = {2025},
 archivePrefix = {arXiv}
}

@inproceedings{IppolitoDCE20,
 author = {Daphne Ippolito and
Daniel Duckworth and
Chris Callison{-}Burch and
Douglas Eck},
 booktitle = "Proceedings of the 58th Annual Meeting of the Association for Computational Linguistics",
 publisher = "Association for Computational Linguistics",
 series = {ACL~'20},
 NOmonth = jul,
 title = {Automatic Detection of Generated Text is Easiest when Humans are Fooled},
 doi = "10.18653/v1/2020.acl-main.164",
 pages = "1808--1822",
 year = {2020}
}

@inproceedings{jiao-etal-2025-rangedetector,
 author = {Jiao, Kaijie  and
Wang, Quan  and
Zhang, Licheng  and
Guo, Zikang  and
Mao, Zhendong},
 booktitle = "Findings of the Association for Computational Linguistics: ACL 2025",
 series = {Findings of ACL~'25},
 NOmonth = jul,
 address = {Vienna, Austria},
 publisher = "Association for Computational Linguistics",
 doi = "10.18653/v1/2025.findings-acl.469",
 pages = "8971--8983",
 ISBN = "979-8-89176-256-5",
 title = {{M}-{R}ange{D}etector: Enhancing Generalization in Machine-Generated Text Detection through Multi-Range Attention Masks},
 year = {2025}
}

@article{jung2025groupadaptivethresholdoptimizationrobust,
 archivePrefix = {arXiv},
 author = {Minseok Jung and Cynthia Fuertes Panizo and Liam Dugan and Yi R. and Fung and Pin-Yu Chen and Paul Pu Liang},
 journal = {ArXiv preprint},
 volume = {arXiv:2502.04528},
 eprint = {2502.04528},
 primaryClass = {cs.CL},
 title = {Group-Adaptive Threshold Optimization for Robust {AI}-Generated Text Detection},
 year = {2025}
}

@inproceedings{karamolegkou2023copyright,
 author = {Antonia Karamolegkou and Jiaang Li and Li Zhou and Anders S{\o}gaard},
 title = {Copyright Violations and Large Language Models},
 year = {2023},
 booktitle = {Proceedings of the 2023 Conference on Empirical Methods in Natural Language Processing},
 series  = {EMNLP~'23},
 NOmonth = dec,
 address = {Singapore},
 publisher = "Association for Computational Linguistics",
 doi = "10.18653/v1/2023.emnlp-main.458",
 pages = "7403--7412"
}

@inproceedings{liu-etal-2025-llms,
 author = {Liu, Jiongnan  and
Zhu, Yutao  and
Wang, Shuting  and
Wei, Xiaochi  and
Min, Erxue  and
Lu, Yu  and
Wang, Shuaiqiang  and
Yin, Dawei  and
Dou, Zhicheng},
 booktitle = {Proceedings of the 63rd Annual Meeting of the Association for Computational Linguistics (Volume 1: Long Papers)},
 series = {ACL~'25},
 NOmonth = jul,
 address = {Vienna, Austria},
 title = {{LLM}s + Persona-Plug = Personalized {LLM}s},
 year = {2025},
 publisher = "Association for Computational Linguistics",
 doi = "10.18653/v1/2025.acl-long.461",
 pages = "9373--9385",
 ISBN = "979-8-89176-251-0"
}

@article{liu2019robertarobustlyoptimizedbert,
 archivePrefix = {arXiv},
 author = {Yinhan Liu and Myle Ott and Naman Goyal and Jingfei Du and Mandar Joshi and Danqi Chen and Omer Levy and Mike Lewis and Luke Zettlemoyer and Veselin Stoyanov},
 journal = {ArXiv preprint},
 volume = {arXiv:1907.11692},
 eprint = {1907.11692},
 primaryClass = {cs.CL},
 title = {{{RoBERTa}: A Robustly Optimized {BERT} Pretraining Approach}},
 year = {2019}
}

@article{llama3,
 author = {Abhimanyu Dubey and Abhinav Jauhri and Abhinav Pandey and Abhishek Kadian and Ahmad Al-Dahle and Aiesha Letman and Akhil Mathur and Alan Schelten and Amy Yang and Angela Fan and Anirudh Goyal and Anthony Hartshorn and Aobo Yang and Archi Mitra and Archie Sravankumar and Artem Korenev and Arthur Hinsvark and Arun Rao and Aston Zhang and Aurélien Rodriguez and Austen Gregerson and Ava Spataru and Baptiste Rozière and Bethany Biron and Binh Tang and Bobbie Chern and Charlotte Caucheteux and Chaya Nayak and Chloe Bi and Chris Marra and Chris McConnell and Christian Keller and Christophe Touret and Chunyang Wu and Corinne Wong and Cristian Canton Ferrer and Cyrus Nikolaidis and Damien Allonsius and Daniel Song and Danielle Pintz and Danny Livshits and David Esiobu and Dhruv Choudhary and Dhruv Mahajan and Diego Garcia-Olano and Diego Perino and Dieuwke Hupkes and Egor Lakomkin and Ehab AlBadawy and Elina Lobanova and Emily Dinan and Eric Michael Smith and Filip Radenovic and Frank Zhang and Gabriel Synnaeve and Gabrielle Lee and Georgia Lewis Anderson and Graeme Nail and Grégoire Mialon and Guan Pang and Guillem Cucurell and Hailey Nguyen and Hannah Korevaar and Hu Xu and Hugo Touvron and Iliyan Zarov and Imanol Arrieta Ibarra and Isabel M. Kloumann and Ishan Misra and Ivan Evtimov and Jade Copet and Jaewon Lee and Jan Geffert and Jana Vranes and Jason Park and Jay Mahadeokar and Jeet Shah and Jelmer van der Linde and Jennifer Billock and Jenny Hong and Jenya Lee and Jeremy Fu and Jianfeng Chi and Jianyu Huang and Jiawen Liu and Jie Wang and Jiecao Yu and Joanna Bitton and Joe Spisak and Jongsoo Park and Joseph Rocca and Joshua Johnstun and Joshua Saxe and Junteng Jia and Kalyan Vasuden Alwala and Kartikeya Upasani and Kate Plawiak and Ke Li and Kenneth Heafield and Kevin Stone and et al.},
 archivePrefix = {arXiv},
 journal = {ArXiv preprint},
 volume = {arXiv:2407.21783},
 eprint = {2407.21783},
 primaryClass = {cs.AI},
 title = {The {Llama} 3 Herd of Models},
 year = {2024}
}

@inproceedings{MackoKMS25,
 author = {Dominik Macko and
Jakub Kopal and
R{\'{o}}bert M{\'{o}}ro and
Ivan Srba},
 booktitle = {Proceedings of the 63rd Annual Meeting of the Association for Computational Linguistics (Volume 1: Long Papers)},
 series = {ACL~'25},
 NOmonth = jul,
 address = {Vienna, Austria},
 title = {{{MultiSocial}: Multilingual Benchmark of Machine-Generated Text Detection
 of Social-Media Texts}},
 year = {2025},
 publisher = "Association for Computational Linguistics",
 doi = "10.18653/v1/2025.acl-long.36",
 pages = "727--752",
 ISBN = "979-8-89176-251-0"
}

@article{mcglinchey2024usingmachinelearningdistinguish,
 archivePrefix = {arXiv},
 author = {Andrea Cristina McGlinchey and Peter J Barclay},
 journal = {ArXiv preprint},
 volume = {arXiv:2412.15253},
 eprint = {2412.15253},
 primaryClass = {cs.CL},
 title = {Using Machine Learning to Distinguish Human-written from Machine-generated Creative Fiction},
 year = {2024}
}

@article{mendoza2024adaptiveselfsupervisedlearningstrategies,
 archivePrefix = {arXiv},
 author = {Rafael Mendoza and Isabella Cruz and Richard Liu and Aarav Deshmukh and David Williams and Jesscia Peng and Rohan Iyer},
 journal = {ArXiv preprint},
 volume = {arXiv:2409.16973},
 eprint = {2409.16973},
 primaryClass = {cs.CL},
 title = {{Adaptive Self-Supervised Learning Strategies for Dynamic On-Device {LLM} Personalization}},
 year = {2024}
}

@inproceedings{mgtbench0001SC0024,
 author = {Xinlei He and
Xinyue Shen and
Zeyuan Chen and
Michael Backes and
Yang Zhang},
 booktitle = {Proceedings of the ACM SIGSAC Conference on Computer and Communications Security},
 series = {CCS~'24},
 address = {Salt Lake City, UT, USA},
 pages = {2251--2265},
 title = {MGTBench: Benchmarking Machine-Generated Text Detection},
 year = {2024}
}

@inproceedings{mikolov2013linguistic,
 author = {Mikolov, Tom{\'a}{\v{s}} and Yih, Wen-tau and Zweig, Geoffrey},
 title = {Linguistic Regularities in Continuous Space Word Representations},
 booktitle = "Proceedings of the 2013 Conference of the North {A}merican Chapter of the Association for Computational Linguistics: Human Language Technologies",
 series = {NAACL~'13},
 NOmonth = jun,
 year = "2013",
 address = "Atlanta, Georgia",
 publisher = "Association for Computational Linguistics",
 pages = "746--751"
}

@inproceedings{mysore-etal-2024-pearl,
 author = {Mysore, Sheshera  and
Lu, Zhuoran  and
Wan, Mengting  and
Yang, Longqi  and
Sarrafzadeh, Bahareh  and
Menezes, Steve  and
Baghaee, Tina  and
Gonzalez, Emmanuel Barajas  and
Neville, Jennifer  and
Safavi, Tara},
 booktitle = {Proceedings of the 1st Workshop on Customizable NLP: Progress and Challenges in Customizing NLP for a Domain, Application, Group, or Individual (CustomNLP4U)},
 title = {Pearl: Personalizing Large Language Model Writing Assistants with Generation-Calibrated Retrievers},
 NOmonth = nov,
 year = {2024},
 address = "Miami, Florida, USA",
 publisher = "Association for Computational Linguistics",
 doi = "10.18653/v1/2024.customnlp4u-1.16",
 pages = "198--219"
}

@inproceedings{narayan-etal-2018-dont,
 author = {Narayan, Shashi  and
Cohen, Shay B.  and
Lapata, Mirella},
 booktitle = {Proceedings of the 2018 Conference on Empirical Methods in Natural Language Processing},
 series = {EMNLP~'18},
 address = {Brussels, Belgium},
 title = {Don{'}t Give Me the Details, Just the Summary! {T}opic-Aware Convolutional Neural Networks for Extreme Summarization},
 year = {2018},
 NOmonth =  oct # "-" # nov,
 publisher = "Association for Computational Linguistics",
 doi = "10.18653/v1/D18-1206",
 pages = "1797--1807"
}

@inproceedings{NEURIPS2020_1457c0d6,
 author = {Brown, Tom and Mann, Benjamin and Ryder, Nick and Subbiah, Melanie and Kaplan, Jared D and Dhariwal, Prafulla and Neelakantan, Arvind and Shyam, Pranav and Sastry, Girish and Askell, Amanda and Agarwal, Sandhini and Herbert-Voss, Ariel and Krueger, Gretchen and Henighan, Tom and Child, Rewon and Ramesh, Aditya and Ziegler, Daniel and Wu, Jeffrey and Winter, Clemens and Hesse, Chris and Chen, Mark and Sigler, Eric and Litwin, Mateusz and Gray, Scott and Chess, Benjamin and Clark, Jack and Berner, Christopher and McCandlish, Sam and Radford, Alec and Sutskever, Ilya and Amodei, Dario},
 booktitle = {Proceedings of the Conference on Neural Information Processing Systems},
 series = {NeurIPS~'20},
 address = {Vancouver, Canada},
 title = {Language Models are Few-Shot Learners},
 year = {2020}
}

@article{quaremba2025wetbenchbenchmarkdetectingtaskspecific,
 archivePrefix = {arXiv},
 author = {Gerrit Quaremba and Elizabeth Black and Denny Vrandečić and Elena Simperl},
 journal = {ArXiv preprint},
 volume = {arXiv:2507.03373},
 eprint = {2507.03373},
 primaryClass = {cs.CL},
 title = {{WETBench}: A Benchmark for Detecting Task-Specific Machine-Generated Text on {Wikipedia}},
 year = {2025}
}

@article{qwen3technicalreport,
 archivePrefix = {arXiv},
 author = {Qwen Team},
 journal = {ArXiv preprint},
 volume = {arXiv:2505.09388},
 eprint = {2505.09388},
 primaryClass = {cs.CL},
 title = {Qwen3 Technical Report},
 year = {2025}
}

@inproceedings{ReimersG19,
 author = {Nils Reimers and
Iryna Gurevych},
 booktitle = {Proceedings of the 2019 Conference on Empirical Methods in Natural Language Processing and the 9th International Joint Conference on Natural Language Processing},
 series = {EMNLP~'19},
 address = {Hong Kong, China},
 pages = {3980--3990},
 title = {{Sentence-BERT}: Sentence Embeddings using {S}iamese {BERT}-Networks},
 year = {2019},
 NOmonth = nov,
 publisher = "Association for Computational Linguistics",
 doi = "10.18653/v1/D19-1410",
 pages = "3982--3992"
}

@inproceedings{russell-etal-2025-people,
 author = {Russell, Jenna  and
Karpinska, Marzena  and
Iyyer, Mohit},
booktitle = {Proceedings of the 63rd Annual Meeting of the Association for Computational Linguistics},
 series = {ACL~'25},
 address = {Vienna, Austria},
 title = {People who frequently use {C}hat{GPT} for writing tasks are accurate and robust detectors of {AI}-generated text},
 year = {2025},
 NOmonth = jul,
 publisher = "Association for Computational Linguistics",
 doi = "10.18653/v1/2025.acl-long.267",
 pages = "5342--5373",
 ISBN = "979-8-89176-251-0"
}

@inproceedings{schler2006effects,
 author = {Schler, Jonathan and Koppel, Moshe and Argamon, Shlomo and Pennebaker, James W},
 booktitle = {Proceedings of the AAAI Conference on Artificial Intelligence},
 series = {AAAI~'06},
 address = {Boston, Massachusetts, USA},
 title = {Effects of age and gender on blogging.},
 year = {2006}
}

@article{Schweinhart21,
 author = {Benjamin Schweinhart},
 journal = {Discrete \& Computational Geometry},
 pages = {331--364},
 title = {Persistent Homology and the Upper Box Dimension},
 year = {2021}
}

@article{shi2024continual,
author = {Shi, Haizhou and Xu, Zihao and Wang, Hengyi and Qin, Weiyi and Wang, Wenyuan and Wang, Yibin and Wang, Zifeng and Ebrahimi, Sayna and Wang, Hao},
title = {Continual Learning of Large Language Models: A Comprehensive Survey},
year = {2025},
issue_date = {April 2026},
publisher = {Association for Computing Machinery},
NOaddress = {New York, NY, USA},
volume = {58},
number = {5},
issn = {0360-0300},
journal = {ACM Computing Surveys},
articleno = {120},
numpages = {42}
}

@article{solaiman2019releasestrategiessocialimpacts,
 archivePrefix = {arXiv},
 author = {Irene Solaiman and Miles Brundage and Jack Clark and Amanda Askell and Ariel Herbert-Voss and Jeff Wu and Alec Radford and Gretchen Krueger and Jong Wook Kim and Sarah Kreps and Miles McCain and Alex Newhouse and Jason Blazakis and Kris McGuffie and Jasmine Wang},
 journal = {ArXiv preprint},
 volume = {arXiv:1908.09203},
 eprint = {1908.09203},
 primaryClass = {cs.CL},
 title = {Release Strategies and the Social Impacts of Language Models},
 year = {2019}
}

@inproceedings{su-etal-2023-detectllm,
 author = {Su, Jinyan  and
Zhuo, Terry  and
Wang, Di  and
Nakov, Preslav},
 booktitle = {Findings of the Association for Computational Linguistics: EMNLP 2023},
 series = {EMNLP~'23},
 address = {Singapore, Singapore},
 title = {{D}etect{LLM}: Leveraging Log Rank Information for Zero-Shot Detection of Machine-Generated Text},
 year = {2023},
 NOmonth = dec,
 publisher = "Association for Computational Linguistics",
 doi = "10.18653/v1/2023.findings-emnlp.827",
 pages = "12395--12412"
}

@inproceedings{su2023detectllm,
 author = {Jinyan Su and Terry Yue Zhuo and Di Wang and Preslav Nakov},
 booktitle = {Proceedings of the 2023 Conference on Empirical Methods in Natural Language Processing},
 series = {EMNLP~'23},
 address = {Singapore, Singapore},
 title = {{D}etect{LLM}: Leveraging Log Rank Information for Zero-Shot Detection of Machine-Generated Text},
 year = {2023},
 NOmonth = dec,
 publisher = "Association for Computational Linguistics",
 doi = "10.18653/v1/2023.findings-emnlp.827",
 pages = "12395--12412"
}

@inproceedings{tian2024multiscale,
 author = {Yuchuan Tian and Hanting Chen and Xutao Wang and Zheyuan Bai and Qinghua Zhang and Ruifeng Li and Chao Xu and Yunhe Wang},
 booktitle = {Proceedings of the Twelfth International Conference on Learning Representations},
 series = {ICLR~'24},
 address = {Vienna, Austria},
 title = {Multiscale Positive-Unlabeled Detection of {AI}-Generated Texts},
 year = {2024}
}

@inproceedings{tian2025symbolic,
 author = {Tian, Chong and Ho, Qirong and Chen, Xiuying},
 booktitle = {Proceedings of the 2025 Conference on Empirical Methods in Natural Language Processing},
 series = {EMNLP~'25},
 address = {Suzhou, China},
 title = {A Symbolic Adversarial Learning Framework for Evolving Fake News Generation and Detection},
 year = {2025},
 NOmonth = nov,
 publisher = "Association for Computational Linguistics",
 doi = "10.18653/v1/2025.emnlp-main.619",
 pages = "12296--12310",
 ISBN = "979-8-89176-332-6"
}

@inproceedings{tseng-etal-2024-two,
 author = {Tseng, Yu-Min  and
Huang, Yu-Chao  and
Hsiao, Teng-Yun  and
Chen, Wei-Lin  and
Huang, Chao-Wei  and
Meng, Yu  and
Chen, Yun-Nung},
 booktitle = {Findings of the Association for Computational Linguistics: EMNLP 2024},
 series = {EMNLP~'24},
 address = {Miami, Florida, USA},
 title = {Two Tales of Persona in {LLM}s: A Survey of Role-Playing and Personalization},
 year = {2024},
 NOmonth = nov,
 publisher = "Association for Computational Linguistics",
 doi = "10.18653/v1/2024.findings-emnlp.969",
 pages = "16612--16631"
}

@inproceedings{tu2024charactereval,
 author = {Tu, Quan and Fan, Shilong and Tian, Zihang and Shen, Tianhao and Shang, Shuo and Gao, Xin and Yan, Rui},
 address = {Bangkok, Thailand},
 booktitle = {Proceedings of the 62nd Annual Meeting of the Association for Computational Linguistics (Volume 1: Long Papers)},
 series = {ACL~'24},
 doi = {10.18653/v1/2024.acl-long.638},
 pages = {11836--11850},
 publisher = {Association for Computational Linguistics},
 title = {{C}haracter{E}val: A Chinese Benchmark for Role-Playing Conversational Agent Evaluation},
 year = {2024}
}

@inproceedings{tulchinskii2023intrinsic,
 author = {Eduard Tulchinskii and Kristian Kuznetsov and Kushnareva Laida and Daniil Cherniavskii and Sergey Nikolenko and Evgeny Burnaev and Serguei Barannikov and Irina Piontkovskaya},
 booktitle = {Proceedings of the Conference on Neural Information Processing Systems},
 series = {NeurIPS~'23},
 address = {New Orleans, Louisiana, USA},
 title = {Intrinsic Dimension Estimation for Robust Detection of {AI}-Generated Texts},
 year = {2023}
}

@inproceedings{wang-etal-2024-m4,
 author = {Wang, Yuxia  and
Mansurov, Jonibek  and
Ivanov, Petar  and
Su, Jinyan  and
Shelmanov, Artem  and
Tsvigun, Akim  and
Whitehouse, Chenxi  and
Mohammed Afzal, Osama  and
Mahmoud, Tarek  and
Sasaki, Toru  and
Arnold, Thomas  and
Aji, Alham Fikri  and
Habash, Nizar  and
Gurevych, Iryna  and
Nakov, Preslav},
 address = {St. Julian's, Malta},
 booktitle = {Proceedings of the 18th Conference of the European Chapter of the Association for Computational Linguistics (Volume 1: Long Papers)},
 series = {EACL~'24},
 doi = {10.18653/v1/2024.eacl-long.83},
 pages = {1369--1407},
 NOpublisher = {Association for Computational Linguistics},
 title = {{{M4}: Multi-generator, Multi-domain, and Multi-lingual Black-Box Machine-Generated Text Detection}},
 year = {2024},
 NOmonth = mar
}

@inproceedings{wang2025beyond,
 author = {Wang, Zixiao and Zhang, Duzhen and Agrawal, Ishita and Gao, Shen and Song, Le and Chen, Xiuying},
 title = {{Beyond Profile: From Surface-Level Facts to Deep Persona Simulation in {LLMs}}},
 year = {2025},
 NOmonth = jul,
 address = {Vienna, Austria},
 booktitle = {Findings of the Association for Computational Linguistics: ACL 2025},
 series = {ACL~'25},
 doi = {10.18653/v1/2025.findings-acl.1094},
 pages = {21239--21257},
 NOpublisher = {Association for Computational Linguistics}
}

@inproceedings{wei2025shortphd,
 author = {Dongjun Wei and Minjia Mao and Xiao Fang and Michael Chau},
 booktitle = {Second Conference on Language Modeling},
 title = {Short-{PHD}: Detecting Short {LLM}-generated Text with Topological Data Analysis After Off-topic Content Insertion},
 series = {COLM~'25},
 address = {Montreal, Canada},
 year = {2025},
}

@inproceedings{wozniak2024personalized,
 author = {Wo{\'z}niak, Stanis{\l}aw and Koptyra, Bart{\l}omiej and Janz, Arkadiusz and Kazienko, Przemys{\l}aw and Koco{\'n}, Jan},
 booktitle = {Proceedings of the 24th IEEE International Conference on Data Mining},
 series = {ICDM~'24},
 address = {Abu Dhabi, UAE},
 title = {Personalized Large Language Models},
 year = {2024}
}

@inproceedings{xu-etal-2025-personalized,
 author = {Xu, Yiyan  and
Zhang, Jinghao  and
Salemi, Alireza  and
Hu, Xinting  and
Wang, Wenjie  and
Feng, Fuli  and
Zamani, Hamed  and
He, Xiangnan  and
Chua, Tat-Seng},
 booktitle = {Proceedings of the 63rd Annual Meeting of the Association for Computational Linguistics},
 series = {ACL~'25},
 address = {Vienna, Austria},
 title = {Personalized Generation In Large Model Era: A Survey},
 year = {2025},
 NOmonth = jul,
 NOpublisher = "Association for Computational Linguistics",
 doi = "10.18653/v1/2025.acl-long.1201",
 pages = "24607--24649",
 ISBN = "979-8-89176-251-0"
}

@inproceedings{xu2025trainingfree,
 author = {Yihuai Xu and Yongwei Wang and Yifei Bi and Huangsen Cao and Zhouhan Lin and Yu Zhao and Fei Wu},
 booktitle = {Proceedings of the Thirteenth International Conference on Learning Representations},
 series = {ICLR~'25},
 address = {Singapore},
 title = {Training-Free {LLM}-generated Text Detection by Mining Token Probability Sequences},
 year = {2025}
}

@inproceedings{zhang-etal-2025-llms,
 author = {Zhang, Denghui  and
Xu, Zhaozhuo  and
Zhao, Weijie},
 booktitle = {Proceedings of the 63rd Annual Meeting of the Association for Computational Linguistics},
 series = {ACL~'25},
 address = {Vienna, Austria},
 title = {{LLM}s and Copyright Risks: Benchmarks and Mitigation Approaches},
 year = {2025},
 NOmonth = may,
 NOpublisher = "Association for Computational Linguistics",
 doi = "10.18653/v1/2025.naacl-tutorial.7",
 pages = "44--50",
 ISBN = "979-8-89176-193-3",
}

@article{herbold2024largelanguagemodelsimpersonate,
 archivePrefix = {arXiv},
 author = {Steffen Herbold and Alexander Trautsch and Zlata Kikteva and Annette Hautli-Janisz},
 journal = {ArXiv preprint},
 volume = {arXiv:2407.12855},
 eprint = {2407.12855},
 primaryClass = {cs.CL},
 title = {Large Language Models can impersonate politicians and other public figures},
 year = {2024}
}

@article{zhang2025personaagentlargelanguagemodel,
 archivePrefix = {arXiv},
 author = {Weizhi Zhang and Xinyang Zhang and Chenwei Zhang and Liangwei Yang and Jingbo Shang and Zhepei Wei and Henry Peng Zou and Zijie Huang and Zhengyang Wang and Yifan Gao and Xiaoman Pan and Lian Xiong and Jingguo Liu and Philip S. Yu and Xian Li},
 journal = {ArXiv preprint},
 volume = {arXiv:2506.06254},
 eprint = {2506.06254},
 primaryClass = {cs.AI},
 title = {{{PersonaAgent}: When Large Language Model Agents Meet Personalization at Test Time}},
 year = {2025}
}

@article{zhang2025personalization,
 author = {Zhehao Zhang and Ryan A. Rossi and Branislav Kveton and Yijia Shao and Diyi Yang and Hamed Zamani and Franck Dernoncourt and Joe Barrow and Tong Yu and Sungchul Kim and Ruiyi Zhang and Jiuxiang Gu and Tyler Derr and Hongjie Chen and Junda Wu and Xiang Chen and Zichao Wang and Subrata Mitra and Nedim Lipka and Nesreen K. Ahmed and Yu Wang},
 journal={Transactions on Machine Learning Research},
 issn={2835-8856},
 year={2025},
 title = {Personalization of Large Language Models: A Survey},
 year = {2025}
}

\newpage
\appendix
\label{sec:appendix}

\section{Details on \dataset and \texttt{M4}}
\label{app:dataset}
\subsection{\textcolor{blue}{Dataset Construction Discussion}}
\label{app:disc-dataset}

\subsubsection{\textcolor{blue}{Scenario Selection}}

\textcolor{blue}{
\texttt{StyloBench} focuses on two personalization scenarios: literary style imitation and social media style blog generation.  
These scenarios are widely used in recent personalization studies~\cite{xu-etal-2025-personalized,mcglinchey2024usingmachinelearningdistinguish} and provide rich individual expression.  
Literary texts offer long narratives with stable author styles, while blog posts capture informal and self-expressive writing.  
Together, they form a practical testbed for personalized MGT, where detectors face realistic stylistic shifts when distinguishing MGT from HWT.  
The current version of \texttt{StyloBench} does not yet include multilingual or other specialized domains, and future extensions are planned to cover a broader range of personalized scenarios.
}

\subsubsection{\textcolor{blue}{Model Selection}}

\textcolor{blue}{
The 2 components of \texttt{StyloBench} use different model configurations, mainly due to differences in data availability and personalization depth.  
For literary data in \texttt{Stylo-Literary}, there are many texts per author, which makes CPT on each author feasible and allows deeper stylistic imitation. Smaller models are used in this setting because they are easier to train and adapt to individual authors.  
For blogs in \texttt{Stylo-Blog}, each user has fewer posts, so CPT is not suitable. In this case, large instruction-tuned models accessed through APIs are prompted to mimic user style, capturing shallower but flexible personalization.  
\texttt{Stylo-Literary} and \texttt{Stylo-Blog} represent two complementary levels of personalization and provide a diverse evaluation space for detectors.
}

\subsubsection{\textcolor{blue}{Dataset Scale and Diversity}}

\textcolor{blue}{
The overall scale of \texttt{StyloBench} is comparable to existing MGT detection benchmarks such as \texttt{CH3-English}~\cite{guo-etal-2023-hc3}, with about $25{,}000$ versus $26{,}000$ samples.  
\texttt{StyloBench} contains 21 subsets, similar to the 20 subsets in \texttt{M4}, and each setting uses over $1{,}000$ test samples, ensuring stable analysis while keeping the benchmark manageable.  

In \texttt{Stylo-Literary}, seven authors with distinct styles are included, all of which lead to clear performance shifts in detectors on HWT and MGT.  
This suggests that the current author set is already informative, while still allowing future expansion.
}

\subsection{Preprocessing for \texttt{Stylo-Literary}}
\label{app:stylo-liter}
Due to formatting and compilation issues, the Gutenberg Book Corpus contains various artifacts that may distort detector performance. These artifacts are removed by eliminating non-original indentation symbols, normalizing line breaks by deleting isolated ones and reducing consecutive breaks to a single instance, and removing lines with repeated ``='' symbols that often mark structural boundaries. 
Unrelated segments are also excluded. We also remove lines with links as they usually indicate source annotations. Lines with long digit sequences are removed since they often correspond to compiler contact information.

\subsection{Details of \texttt{Stylo-Blog}}
\label{app:stylo-blog}

\paragraph{Data Source.}
The data source of \texttt{Stylo-Blog} is Blog-1K, a well-known subset of the Blog Authorship Corpus~\cite{schler2006effects}. The Blog Authorship Corpus was originally collected from posts on \texttt{blogger.com} in 2004. It spans a wide range of topics, is linguistically diverse, and has been widely used for studying stylistic and demographic features.  
Based on this corpus, Blog-1K filters high-quality blogs and groups them by authors, resulting in 1,000 authors and 16,132 posts. It provides clearer author attribution, a more balanced data distribution, and more consistent writing quality, making it suitable for constructing \texttt{Stylo-Blog} in personalized evaluation settings.

\begin{figure*}[!t]
\centering
\includegraphics[width=\columnwidth]{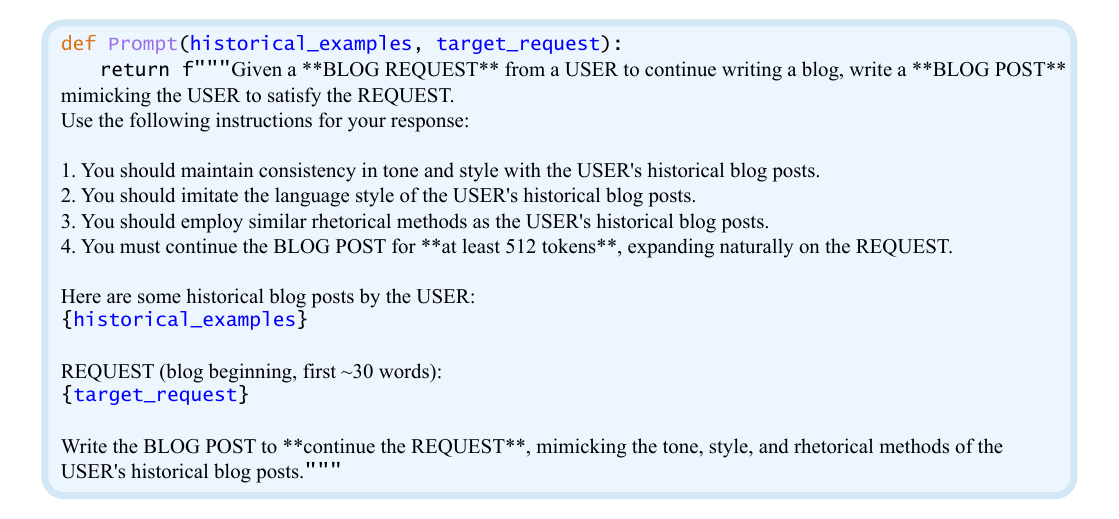} 
\caption{
Prompt template for MGT in synthesizing \texttt{Stylo-Blog}. It takes 1–3 blogs from the same author as \texttt{historical\_examples} and uses the first 30 tokens of the current blog as \texttt{target\_request} for continuation.
}
\label{fig:prompt}
\end{figure*}

\paragraph{Few-shot Construction.}
For each author, we construct few-shot prompts using their remaining posts to guide style imitation.
After selecting one post as the target for generating MGT, the number of few-shot examples depends on how many posts remain for that author.
If only one post remains, we use a 1-shot setting.
If two posts remain, we use a 2-shot setting.
If three or more posts remain, we randomly sample three posts to form a 3-shot prompt.
This strategy ensures that the few-shot construction adapts to data availability while maintaining consistency across authors. The prompt template is shown in Figure~\ref{fig:prompt}.

\subsection{Configurations of LLM Generators}
\label{app:llm-config}

\begin{table}[!t]
\centering
\small
\setlength{\tabcolsep}{8pt}
\begin{tabular}{ll}
\toprule
\textbf{Parameter} & \textbf{Value} \\
\midrule
Maximum samples & 3,000 \\
Batch size & 8 \\
Learning rate & 0.0001 \\
Epochs & 5 \\
\midrule
\multicolumn{2}{l}{\textit{LoRA configuration}} \\
\midrule
LoRA rank & 16 \\
LoRA alpha & 32 \\
LoRA dropout & 0.1 \\
\bottomrule
\end{tabular}
\caption{Key hyperparameters in CPT for generating \texttt{Stylo-Literary} MGTs.}
\label{tab:train_params}
\end{table}

In constructing \texttt{Stylo-Literary}, key parameters are shown in Table~\ref{tab:train_params}. 
For both subsets, we set the maximum length to 512 tokens for consistent style imitation and use a temperature of 1 to encourage diversity, while keeping other parameters at their default values.

\subsection{The English Subset of \texttt{M4}} 
\label{app:m4}
The selected English subset contains 5 data sources to reflect diversity and daily language use: Reddit, Wikipedia, WikiHow, arXiv, and PeerRead. Each LLM generator uses 2–8 distinct prompts to produce varied MGTs. We select only the generators that have data in all five sources in the released version of \texttt{M4}. Table~\ref{tab:m4} reports the detailed statistics.

\begin{table}[!t]
\centering
\resizebox{\textwidth}{!}{
\setlength{\tabcolsep}{4pt}
\begin{tabular}{lcccccc}
\toprule
\textbf{Generator} 
  & \textbf{Wikipedia} & \textbf{Reddit} & \textbf{WikiHow} & \textbf{PeerRead} & \textbf{arXiv} & \textbf{Total}\\
\midrule
Davinci & 3,000 & 3,000 & 3,000 & 2,323 & 3,000 & 14,323\\
ChatGPT & 2,995 & 3,000 & 3,000 & 2,344 & 3,000 & 14,339\\
Cohere  & 2,336 & 1,220 & 2,999 & 1,702 & 3,000 & 11,257\\
BLOOMz  & 3,000 & 3,000 & 3,000 & 2,340 & 3,000 & 14,340\\
\bottomrule
\end{tabular}
}
\caption{Statistics of the selected English subset of \texttt{M4}, where the values indicate the numbers of HWTs and MGTs, which are equal.
}
\label{tab:m4}
\end{table}

\section{Experiment Details}
\label{app:exp}
\subsection{Baselines}
\label{app:baseline}
We follow the same set of baseline detectors as in~\cite{xu2025trainingfree}.  
The implementation details of the seven baselines in this study are as follows.  


\paragraph{Log-Likelihood~\cite{solaiman2019releasestrategiessocialimpacts}.} The average log probability of all tokens in a text is used as the metric. Texts with lower average likelihood are more likely to be MGT.  

\paragraph{LogRank~\cite{solaiman2019releasestrategiessocialimpacts}.} The average log rank of tokens in the text, where ranks are determined by GPT-J’s probabilities, is used as the metric. Texts with higher ranks could be MGT.

\paragraph{Entropy~\cite{gehrmann-etal-2019-gltr}.} The average entropy of the predicted token distribution is computed. Texts with higher entropy values are more likely to be MGT.  

\paragraph{DetectLRR~\cite{su-etal-2023-detectllm}.} The ratio of log-likelihood to log-rank is taken as the score. Larger ratios indicate a higher chance of MGT.  

\paragraph{Fast-DetectGPT~\cite{bao2024fastdetectgpt}.} It perturbs the input text to create contrast samples and compares the scoring differences between the original and perturbed texts. Larger discrepancies indicate that the original text is more likely to be MGT.  

\paragraph{Lastde~\cite{xu2025trainingfree}.} This method analyzes local and global diversity of token probability sequences, and combines them with likelihood information. Lower diversity relative to likelihood is more indicative of MGT.  

\paragraph{Lastde++~\cite{xu2025trainingfree}.} It is an enhanced version of Lastde that normalizes scores using multiple contrast samples, making the results more stable. Higher normalized values suggest MGT.

We test all baselines under a black-box scenario without generator access, using GPT-J-6B~\cite{gpt-j} as a proxy. For Log-Likelihood, LogRank, Entropy, and DetectLRR, features are computed from GPT-J-6B probabilities and ranks; for DetectGPT, Fast-DetectGPT, Lastde, and Lastde++, perturbed variants are generated and scores are compared with the original.

\subsection{\textcolor{blue}{Training-free Methods}}

\textcolor{blue}{Since full results on \texttt{Stylo-Blog} have already been presented in Table~\ref{tab:avg-res}, the complete results for \texttt{Stylo-Literary} and \texttt{M4} are provided in Appendix~\ref{app:full-res} for completeness.

}

\subsection{\textcolor{blue}{Experiments on Training-based Methods}}
\label{app:exp-train}

\paragraph{\textcolor{blue}{Setup.}}
\textcolor{blue}{
We evaluate two representative training-based detectors: RoBERTa~\cite{liu2019robertarobustlyoptimizedbert} and DeTeCtive~\cite{guo2024detective}.  
For \texttt{M4}, each subset is split by randomly selecting $1{,}000$ HWT and MGT samples as the test set, with all remaining samples used for training.  
The finetuning configuration for RoBERTa is shown in Table~\ref{tab:roberta-set}.  
For DeTeCtive, we use the publicly released checkpoint that achieves the best performance on \texttt{M4}\footnote{\url{https://huggingface.co/heyongxin233/DeTeCtive/blob/main/M4_monolingual_best.pth}}
and evaluate it on the same test sets.  
We use AUROC for evaluation.
}

\begin{table}[htbp]
\small
\centering
\textcolor{blue}{
\begin{tabular}{lc}
\toprule
\textbf{Hyperparameter} & \textbf{Value} \\
\midrule
Batch size      & 32 \\
Epochs          & 10 \\
Learning rate   & $2 \times 10^{-5}$ \\
Warmup steps    & 2000 \\
Random seed     & 42 \\
\bottomrule
\end{tabular}
}
\caption{\textcolor{blue}{Training configuration for RoBERTa.}}
\label{tab:roberta-set}
\end{table}

\paragraph{\textcolor{blue}{Results and Analysis.}}
\textcolor{blue}{
The averaged results on the three datasets are reported in Table~\ref{tab:avg-tr}.  
Both detectors perform well on \texttt{M4}, but their AUROC scores on \texttt{StyloBench} mostly fall within $0.4$–$0.6$, close to random prediction in many cases.  
RoBERTa shows near-random performance in about $64\%$ of personalized settings, while this rises to $92\%$ for DeTeCtive, indicating frequent and structural performance drops beyond typical domain generalization settings.

At the same time, these training-based models do not collapse to extremely low AUROC values as some training-free detectors do, indicating that they may rely on a broader set of learned features that help prevent complete failure.
Complete results are provided in Appendix~\ref{app:full-res} for reference.

}
\begin{table}[htbp]
\small
\centering
\textcolor{blue}{
\begin{tabular}{lcc}
\toprule
\textbf{Dataset} & \textbf{RoBERTa} & \textbf{DeTeCtive} \\
\midrule
\texttt{Stylo-Literary} & 53.33 & 53.78 \\
\texttt{Stylo-Blog}     & 64.77 & 49.73 \\
\texttt{M4}             & 99.94 & 84.69 \\
\midrule
\textbf{AUROC 0.4--0.6} & 64\%  & 92\%  \\
\bottomrule
\end{tabular}
}
\caption{\textcolor{blue}{Average AUROC of training-based detectors.}}
\label{tab:avg-tr}
\end{table}

\section{Metrics and Mathematical Tools}
\label{app:math}
\subsection{AUROC}
\label{app:auroc}
AUROC measures detector performance across the full range of thresholds rather than relying on a fixed one, therefore reflecting the overall ability of the detector. AUROC ranges from 0.0 to 1.0, where 0.5 corresponds to random guessing, and 1.0 indicates perfect discrimination. It can be interpreted as the probability that a randomly selected machine-generated text is assigned a higher detection score than a randomly selected human-written text. Values below 0.5 indicate performance worse than random guessing and imply that the predictions are systematically inverted.

\begin{figure*}[!t]
\centering
\includegraphics[width=\columnwidth]{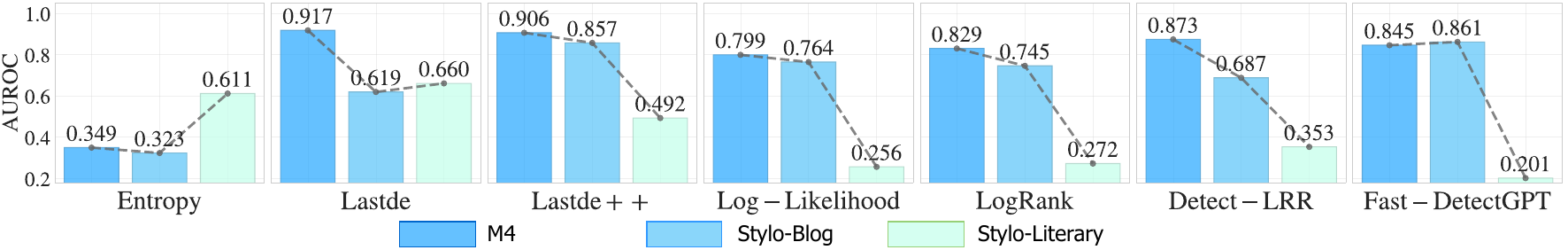} 
\caption{
The average AUROC of different detectors on \texttt{M4}, \texttt{Stylo-Blog}, and \texttt{Stylo-Literary}. The lines indicate performance changes caused by domain transfer. Detectors show a clear change in the personalized domain, including surges, decreases, and inversions.
}
\label{fig:det_tr}
\end{figure*}
\begin{figure*}[!t]
\centering
\includegraphics[width=\columnwidth]{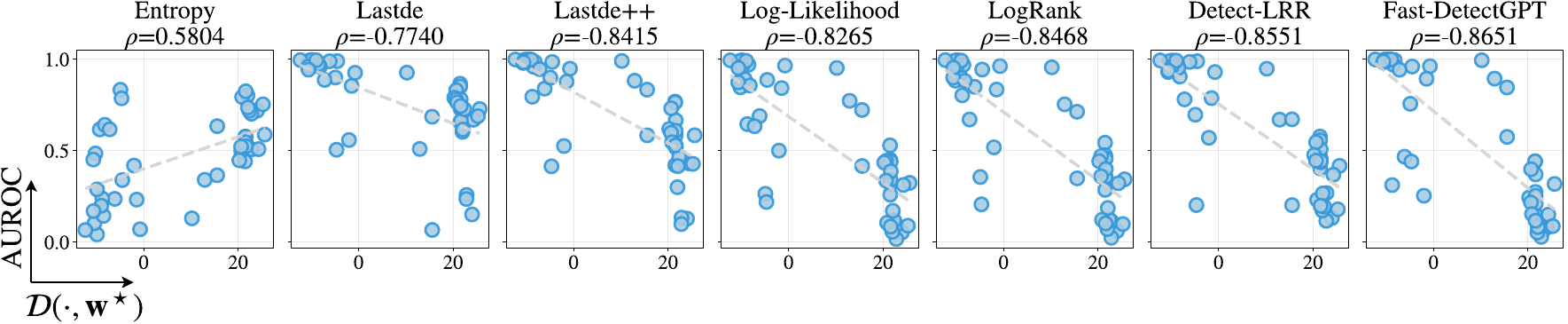} 
\caption{
AUROC of each detector and the distribution of feature value difference \(\mathcal{D}(\cdot,\mathbf{w}^\star)\) on the test set for each experiment. The plots show clear correlation, with varying strength and sign across detectors.
}
\label{fig:corr_viz}
\end{figure*}

\subsection{Correlation Coefficients}
\label{app:corr-coef}

\paragraph{Spearman \(\rho\)} Spearman’s rank correlation coefficient measures the monotonic relationship between two variables by computing the Pearson correlation on their rank values. It is defined as  
\begin{equation}
\rho = 1 - \frac{6 \sum_{i=1}^{n} d_i^2}{n(n^2-1)},    
\end{equation}
where \(d_i\) is the rank difference of the \(i\)-th sample. Values close to 1 or \(-1\) indicate strong positive or negative monotonic correlation.  

\paragraph{Pearson \(r\)} Pearson’s correlation coefficient captures the linear relationship between two variables. It is defined as  
\begin{equation}
\scalebox{1.2}{$
   r = \frac{\sum_{i=1}^{n} (x_i - \bar{x})(y_i - \bar{y})}{\sqrt{\sum_{i=1}^{n}(x_i - \bar{x})^2}\sqrt{\sum_{i=1}^{n}(y_i - \bar{y})^2}}.
$}
\end{equation}
Values near 1 indicate strong positive linear correlation, values near \(-1\) strong negative correlation, and 0 denotes no linear correlation.  

\paragraph{Kendall’s \(\tau\)} Kendall’s \(\tau\) measures the ordinal association between two variables based on the number of concordant and discordant pairs:  
\begin{equation}
\scalebox{1.2}{$
   \tau = \frac{(\text{\#concordant pairs}) - (\text{\#discordant pairs})}{\tbinom{n}{2}}.
$}
\end{equation}
In this work, \(\tau\) is also employed as a quantitative measure of word order. Values close to 1 indicate strong agreement with the original order, values close to \(-1\) indicate reversed order, and values around 0 correspond to randomized tokens.

\subsection{\textcolor{blue}{Mathematical Design Explanation}}
\label{app:rayleigh}
\paragraph{Derivation of Equation~\ref{eqi}}
The equation $q_i(\mathbf{w})=(\mathbf{w}^\top v_G)(\mathbf{w}^\top v_S)$ follows from basic matrix multiplication:
\begin{equation*}
   \resizebox{0.5\textwidth}{!}{
    $(\mathbf{w}^\top v_G)(\mathbf{w}^\top v_S)= (\mathbf{w}^\top v_G)(v_S^\top \mathbf{w}) = \mathbf{w}^\top (v_G v_S^\top)\mathbf{w}.$
} 
\end{equation*}

\paragraph{Derivation of Equation~\ref{get_a}}
For any real matrix $\mathrm{M}\in\mathbb{R}^{n\times n}$ and vector $\mathbf{w}$, the quadratic form equals its transpose, as it is a scalar:
\[
(\mathbf{w}^\top M \mathbf{w})^\top
= \mathbf{w}^\top M^\top (\mathbf{w}^\top){}^\top
= \mathbf{w}^\top M^\top \mathbf{w}.
\]
Therefore
\begin{align}
\mathbf{w}^\top \mathrm{M}\mathbf{w}
&= \textstyle{\tfrac{1}{2}}\mathbf{w}^\top (\mathrm{M}+\mathrm{M}^\top)\mathbf{w}. \nonumber
\end{align}
Applying this to $\mathrm{M}=v_G v_S^\top$ gives
\[
q_i(\mathbf{w})
= \mathbf{w}^\top \tfrac{1}{2}(v_G v_S^\top + v_S v_G^\top)\mathbf{w}.
\]

\paragraph{Rayleigh Quotient} For a matrix \(\mathrm{M} \in \mathbb{R}^{n \times n}\) and a nonzero vector \(w \in \mathbb{R}^n\), the Rayleigh quotient is defined as  
\begin{equation}
   R(w) = \frac{w^\top \mathrm{M}  w}{w^\top w}. 
\end{equation}

When \(||w||=1\), this can be also written as: \(R(w) = {w^\top \mathrm{M}  w}\). When \(\mathrm{M} \) is symmetric,  \(R(w)\) is bounded between the smallest and largest eigenvalues of \(\mathrm{M} \). In particular, the minimum of \(R(w)\) equals the smallest eigenvalue and the maximum equals the largest eigenvalue, attained when \(w\) is the corresponding eigenvector.

\section{Analysis on Feature-Inversion Trap}

\subsection{Performance Change of Detectors}
\label{app:perf-change}
To examine detector behavior across domains, we compute the average AUROC on \texttt{M4}, \texttt{Stylo-Blog}, and \texttt{Stylo-Literary}, and report the results in Figure~\ref{fig:det_tr}. The trends are consistent with~\S\ref{main_res}.

Most detectors degrade more in the personalized domain than in the general domain, while a few, such as Entropy and Fast-DetectGPT, improve on certain personalized subsets. Classifiers show smaller fluctuations on \texttt{Stylo-Blog} than on \texttt{Stylo-Literary}, suggesting that post-training imitation misleads detectors more than prompting. On \texttt{Stylo-Literary}, five detectors fall below 0.5 AUROC, indicating prediction reversal in the personalized domain.

\subsection{Correlation Visualization}
\label{app:corr-viz}
Following \S\ref{corr}, we plot the AUROC of each detector and the distribution of $\mathcal{D}(\cdot,\mathbf{w}^\star)$ on the test set across multiple experiments, as shown in Figure~\ref{fig:corr_viz}. The results reveal that: (1) The sample distribution is concentrated, and most points lie near the fitted line, which indicates a clear correlation between AUROC and $\mathcal{D}(\cdot,\mathbf{w}^\star)$. (2) Entropy shows a positive correlation, while the others show a negative correlation, which indicates that the Feature-Inversion Trap affects different detectors in different ways.

\subsection{\textcolor{blue}{Verification of Detector Dependence on Inverted Features}}
\label{app:dep_verif}
\textcolor{blue}{
In ~\S\ref{corr}, we show a strong negative correlation between the projection gap of inverted features and detector performance. This correlation may still be spurious. We therefore design experiments to show that detectors do rely on inverted features. 

The core idea is simple: if a detector relies on inverted features, then it should still work on random text. The text may be meaningless, but if positive and negative samples show a clear gap along the inverted feature, the detector should still show strong results. On the contrary, other features or pure random noise should not produce this effect.
}

\paragraph{\textcolor{blue}{Setup.}}  
\textcolor{blue}{
We tokenize each text. We then shuffle these tokens to create text with no meaning. Based on these texts, we build three types of test sets. 
(1) \textit{Along inverted direction}. We sort samples by their projection on the inverted feature direction $\mathbf{w}^\star$. We take the top samples as positives and the bottom samples as negatives. This set tests whether the detector depends only on inverted features. 
(2) \textit{Along orthogonal direction}. We sort samples by their projection on a direction $\mathbf{w}^\star_\bot$ that is orthogonal to the inverted feature. We take the top samples as positives and the bottom samples as negatives. This set tests whether the detector depends on other directions that share the same form but do not reflect inverted features. 
(3) \textit{Random}. We shuffle all samples and split them into positives and negatives at random. This gives a baseline under complete randomness. 
For each type, we create 50 datasets. Each has 50 positive and 50 negative samples.
}

\begin{figure*}[!t]
\centering
\includegraphics[width=\columnwidth]{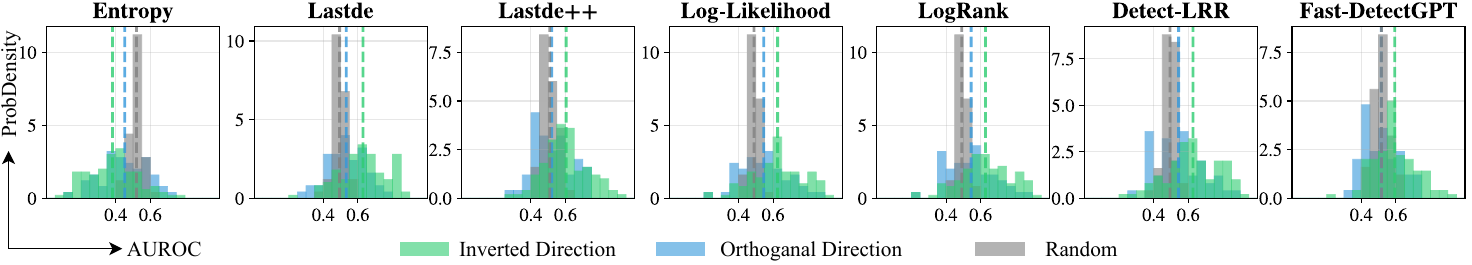} 
\caption{
\textcolor{blue}{AUROC of each detector on three synthetic test sets and the distribution of feature value difference $\mathcal{D}(\cdot,\mathbf{w}^\star)$ under each construction. }
}
\label{fig:pert_distri}
\end{figure*}

\paragraph{\textcolor{blue}{Results and analysis.}}  
\textcolor{blue}{
We test seven detectors on these datasets. We plot the AUROC distribution across the 50 runs in Figure \ref{fig:pert_distri}. The results show three clear patterns. 

(1) \textit{Random}: All detectors stay near 0.5. This confirms that detectors behave randomly in random noise, which confirms their normal functioning.

(2) \textit{Inverted direction}: Detectors show strong classification tendencies. Entropy has an average AUROC below 0.4. Other detectors have an AUROC above 0.6. The AUROC distribution is also more spread out. Extreme values such as AUROC above 0.8 or below 0.2 appear more often. These signs show that detectors rely on inverted features to make a prediction. 

(3) \textit{Orthogonal direction}: No strong tendency appears. Some detectors behave almost the same as in the random set, such as Lastde, Lastde++, and Fast-DetectGPT. Other detectors shift slightly but remain within the 0.4 to 0.6 range, which is still close to random. These results show that other features cannot influence detector behavior as inverted features do. 
Together with the strong correlation shown in Figure \ref{fig:corr-distri}, these results support a clear conclusion. In personalized settings, the feature-inversion trap is a key reason for the reversed behavior of many detectors.
}

\subsection{\textcolor{blue}{Ablation Study for \method}}  
\label{app:abl}
\begin{figure}[!t]
\centering
\includegraphics[width=\columnwidth]{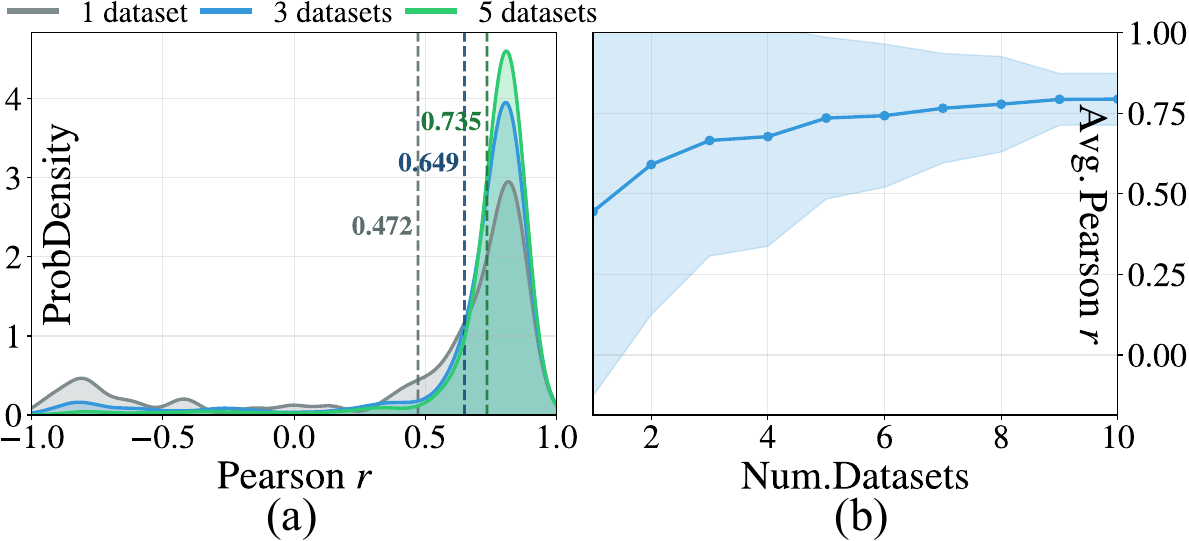} 
\caption{
\textcolor{blue}{(a) Distribution of Pearson \(r\) with 1, 3, and 5 probe datasets.  
(b) Change of mean Pearson \(r\) with the number of probe datasets, shaded with one standard deviation intervals.}
}
\label{fig:multi_scramble_corr}
\end{figure}


\textcolor{blue}{We conduct an ablation study to assess how the number of probe datasets affects the reliability of \method.  
Figure~\ref{fig:multi_scramble_corr}(a) shows the distribution of  \(r\) when using 1, 3, and 5 datasets. With fewer datasets, the mean \(r\) decreases and the probability of \(r<-0.5\) increases, indicating a higher risk of incorrect prediction. Figure~\ref{fig:multi_scramble_corr}(b) plots the change of \(r\) as the number of datasets grows from 1 to 10. The mean \(r\) rises gradually, but slows down as it approaches an upper bound near 0.8. The standard deviation also decreases, but with diminishing returns. These results suggest that increasing the dataset size improves reliability, but the benefit diminishes over time. Using five probe datasets offers a good balance in practice.}

\section{\textcolor{blue}{Empirical Evidence for \S \ref{disc}}}
\label{app:evd}
\subsection{\textcolor{blue}{Difference Between the Feature-Inversion Trap and OOD Effects}}

\label{app:ood-exp}
\textcolor{blue}{
In \S\ref{disc}, we state two reasons why the feature-inversion trap is different from standard OOD effects:  
(1) it is tightly correlated with the inverted feature direction;  
(2) it causes reversal of detector behavior, not only performance decay.  
We now verify both points with experiments.}

\begin{figure}[!t]
\centering
\includegraphics[width=0.8\columnwidth]{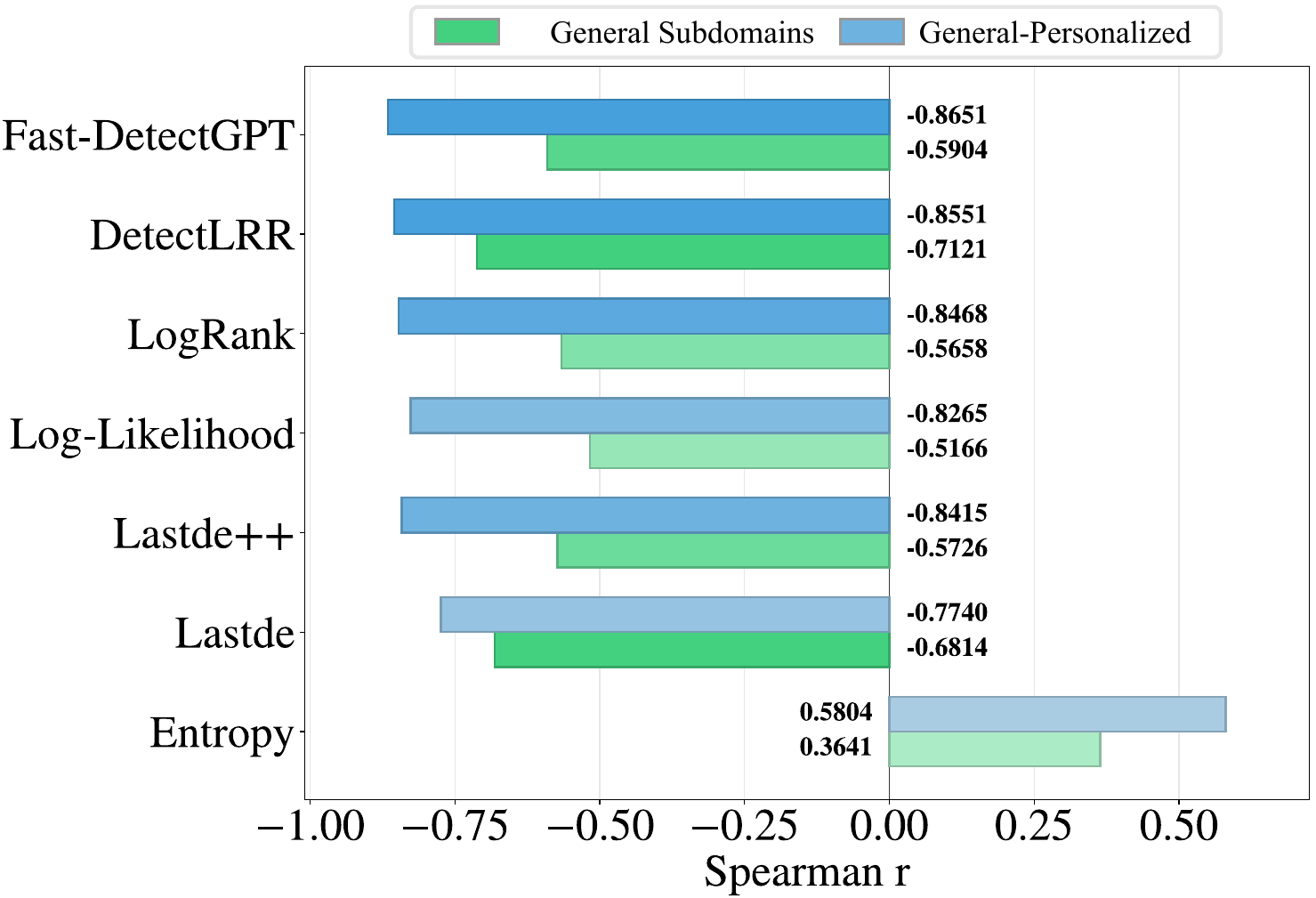} 
\caption{
\textcolor{blue}{Correlation differences in different detectors.}
}
\label{fig:crs-dom-rel}
\end{figure}

\paragraph{\textcolor{blue}{For point (1).}}  
\textcolor{blue}{We use performance differences across M4 subsets as a representative OOD case, as prior work~\cite{wang-etal-2024-m4} studies detector generalization on them and their domain gaps are well known. For all M4 subsets, we follow~\S\ref{verify} to compute the projection gap on the inverted direction, denoted as \(\mathcal{D}_{\mathrm{M4}}\). We then compute the Spearman correlation between this gap and detector performance on M4:
\begin{equation}
    r = \mathrm{Spearman}(\mathcal{D}_{\mathrm{M4}}, \mathrm{AUROC}_{\mathrm{M4}}).
\end{equation}
Results for all detectors are shown in Figure~\ref{fig:crs-dom-rel}. Cross-domain correlations (general~$\rightarrow$~personalized) exceed those within M4, showing that the inverted direction explains shifts to personalized domains but not differences within M4.

}
\paragraph{\textcolor{blue}{For point (2).}}  
\textcolor{blue}{Reversal rarely appears across M4 subsets but is frequent on StyloBench. Complete M4 results are in Appendix~\ref{app:full-res}. Among 140 runs, 25 (17.9\%) have AUROC below 0.5, while in Stylo-Literary, 93 of 147 (63.3\%) fall below 0.5. This gap shows that personalized scenarios cannot be explained by common OOD factors such as feature weakening. Together with point (1), this supports that personalization causes structural feature reversal rather than a typical domain shift.
}

\subsection{\textcolor{blue}{Connections to Other Terms}}
\label{app:con-terms}

\textcolor{blue}{
The feature inversion phenomenon introduced appears to share surface similarities with several established concepts in machine learning.  
To avoid misunderstanding, we describe the relation between these concepts and our findings, and we clarify the differences.
}

\textbf{\textcolor{blue}{Spurious Correlations.}}
\textcolor{blue}{
The feature inversion phenomenon relates to spurious correlations~\cite{abs-2402-12715}, as detectors rely on cues that are not stable across domains.
For example, they often assume HWT is more diverse than MGT, but personalization flips this pattern and changes the cue’s meaning.
Unlike typical spurious correlations, the relation is reversed rather than weakened.
}

\textbf{\textcolor{blue}{Domain Shift Robustness.}}
\textcolor{blue}{
The feature inversion phenomenon is related to domain shift.
In typical settings, performance degrades as features lose reliability~\cite{ZhouLQXL23}.
Under personalization, the feature direction changes, causing prediction reversal.
This indicates a stronger, more structural shift than standard OOD cases, as discussed in Appendix~\ref{app:ood-exp}.
}

\textbf{\textcolor{blue}{Adversarial Robustness.}}
\textcolor{blue}{
For the feature-inversion trap, there is a conceptual link to adversarial robustness, as both involve shifts that redirect model attention to different features~\cite{IlyasSTETM19}.
Personalization consistently alters stylistic properties, changing how detectors use cues.
However, this shift arises naturally from personalized text generation rather than intentional adversarial manipulation, so the mechanisms differ.
}

\textbf{\textcolor{blue}{Invariant Representation Learning.}}
\textcolor{blue}{
Invariant representation learning aims to identify features stable across domains~\cite{irl}.
Feature inversion shows that many detectors rely on non-invariant features, such as lexical or semantic diversity.
Even with invariant learning, detectors may still depend on unstable cues under strong personalization.
This highlights a limitation of invariant learning for MGT detection.
}

\textbf{\textcolor{blue}{Feature Inversion in Natural Language Processing.}}
\textcolor{blue}{
The term feature inversion in NLP often refers to reconstructing text from representations~\cite{morris-etal-2023-text}.
Our use differs: it denotes a change in feature direction across domains.
A feature separating HWT from MGT in one domain may work oppositely in another.
This reflects semantic reversal rather than reconstruction.

}

\subsection{\textcolor{blue}{What Do Inverted Features Capture}}
\label{app:what-inv-feat}

\textcolor{blue}{
We believe that the inverted features include, but are not limited to, lexical diversity and semantic coherence.  
Both of these properties describe aspects of textual coherence, and both show a clear reversal trend across domains.}

\paragraph{\textcolor{blue}{Lexical Diversity}}
\textcolor{blue}{We follow intrinsic dimension analysis from prior work~\cite{tulchinskii2023intrinsic}, which shows MGT embeddings lie on lower-dimensional manifolds than HWT, indicating lower diversity.
Using the probe datasets in Section~4.1, we follow~\cite{tulchinskii2023intrinsic}: \texttt{roberta-base} encodes each sentence, last-layer token embeddings are treated as points, and persistent homology dimension~\cite{Schweinhart21} estimates lexical variation.
Results are in Table~\ref{tab:tokens}.
For HWT, personalized samples have higher diversity than general ones.
For MGT, diversity is lower in general but higher in personalized settings.
This reversal aligns with feature inversion and suggests that in highly personalized scenarios, MGT may use words more variably than HWT.

}
\begin{table}[htbp]
\small
\centering
\textcolor{blue}{
\begin{tabular}{lccc}
\toprule
\makecell{\textbf{Token} \\ \textbf{Dimension}} 
& \textbf{General} 
& \textbf{Personalized} 
& \boldmath$\Delta$ \\
\midrule
HWT & 9.9952 & 12.0255 & 2.0304 \\
MGT & 6.5751 & 13.0849 & 6.5098 \\
\midrule
\boldmath$\Delta$ & -3.4201 & 1.0594 & - \\
\bottomrule
\end{tabular}
}
\caption{\textcolor{blue}{Token-level intrinsic dimension for general and personalized domains.}}
\label{tab:tokens}
\end{table}

\paragraph{\textcolor{blue}{Semantic Coherence}}
\textcolor{blue}{We apply a similar approach to analyze sentence-level semantic coherence.
Using Sentence-BERT~\cite{ReimersG19}, each sentence is mapped to a vector, where closer vectors indicate stronger semantic consistency.
If these vectors lie on a low-dimensional manifold, the text is more coherent with smoother semantic flow, while higher dimensionality indicates larger semantic shifts.
We segment each text into sentences using punctuation and encode them with the \texttt{all-MiniLM-L6-v2} model, then estimate the manifold dimension for each group.
Results are presented in Table~\ref{tab:sent}.
For MGT, coherence is lower in general domains, but in personalized domains it becomes higher than that of HWT.
Therefore, inverted features are related to both lexical diversity and semantic coherence.

}
\begin{table}[!t]
\small
\centering
\textcolor{blue}{
\begin{tabular}{lccc}
\toprule
\makecell{\textbf{Sentence} \\ \textbf{Dimension}}
& \textbf{General} 
& \textbf{Personalized} 
& \boldmath$\Delta$ \\
\midrule
HWT & 5.2764 & 4.4390 & -0.8374 \\
MGT & 3.2783 & 5.4325 & 2.1542 \\
\midrule
\boldmath$\Delta$ & -1.9981 & 0.9935 & - \\
\bottomrule
\end{tabular}
}
\caption{\textcolor{blue}{Sentence-level intrinsic dimension for general and personalized domains.}}
\label{tab:sent}
\end{table}

\subsection{\textcolor{blue}{Towards Mitigating the Feature Inversion Phenomenon}}
\label{app:mitigate}
\textcolor{blue}{
Finetuning on personalized text can improve detector accuracy in the target domain, but these gains do not transfer well across domains. Training on one domain brings little benefit to another, and joint training still cannot overcome the model’s generalization limits. Thus, training alone cannot resolve feature inversion; effective solutions require capturing inverted feature behavior or avoiding reliance on unstable cues.

}

\textcolor{blue}{
We evaluate three finetuning variants of RoBERTa.
For \texttt{M4}, following Appendix~\ref{app:exp-train}, each subset is split with $1{,}000$ samples for testing and the rest for training; \dataset is split similarly.
The first variant is trained on M4, the second on \texttt{Stylo-Literary}, and the third on the combined sets.
All variants are tested on the M4 test sets and both parts of \dataset.
Results are shown in Table~\ref{tab:roberta}.
Finetuning improves performance within the training domain, but gains do not transfer across domains.
Even joint training does not close the gap, especially on \texttt{Stylo-Blog}.
This indicates that limited generalization still restricts mitigation of the inversion phenomenon.

}

\begin{table}[!htbp]
\small
\centering
\textcolor{blue}{
\begin{tabular}{lccc}
\toprule
\textbf{Training Set} & \textbf{Stylo-Literary} & \textbf{Stylo-Blog} & \textbf{M4} \\
\midrule
None & 52.37 & 53.87 & 41.68 \\
M4 & 53.33 & 64.77 & 99.94 \\
Stylo-Literary & 99.96 & 67.95 & 70.72 \\
Mixed & 99.95 & 71.94 & 99.95 \\
\bottomrule
\end{tabular}
}
\caption{\textcolor{blue}{Performance of RoBERTa variants, trained on different data.}}
\label{tab:roberta}
\end{table}







\section{\textcolor{blue}{Full Results}}
\label{app:full-res}

\textcolor{blue}{For training-free methods, full results on \texttt{Stylo-Literary} and \texttt{M4} are shown in Tables~\ref{tab:full_liter} and~\ref{tab:full_m4}. Since the full results on \texttt{Stylo-Blog} are already given in Table~\ref{tab:avg-res}, we do not repeat them here. For training-based methods, full results on \texttt{Stylo-Literary}, \texttt{Stylo-Blog}, and the English subset of \texttt{M4} are shown in Tables~\ref{tab:mgt---tr-ft}, \ref{tab:mgt---tr-pr}, and~\ref{tab:mgt---tr-m4}.}

\begin{table*}[htbp]
    \centering\small
    \begin{tabular}{lcccccccc}
    \midrule
    \diagbox[dir=SE,height=2\line]{\textbf{Detector}}{\textbf{Author}} & \textbf{J.A} & \textbf{C.D} & \textbf{F.D} & \textbf{P.L} & \textbf{B.S} & \textbf{J.S} & \textbf{M.T} & \textbf{Avg.} \\
    \midrule
    \rowcolor{lightgray}
    \multicolumn{9}{c}{{\bf \textit{Llama-3.1-8B}}} \\
    \midrule
    Entropy & 71.91 & 57.64 & 44.19 & 58.80 & 52.07 & 52.12 & 49.85 & 55.23 \\
    Lastde & 15.17 & 72.96 & 86.64 & 72.80 & 85.08 & 73.41 & 83.10 & 69.88 \\
    Lastde++ & 12.85 & 58.34 & 76.84 & 58.45 & 76.40 & 66.63 & 73.17 & 60.38 \\
    Log-Likelihood & 5.33 & 31.81 & 52.93 & 32.42 & 45.10 & 44.07 & 44.47 & 36.59 \\
    LogRank & 6.07 & 34.62 & 54.38 & 34.35 & 47.56 & 45.67 & 47.08 & 38.53 \\
    Detect-LRR & 13.23 & 46.21 & 57.74 & 41.72 & 55.74 & 50.79 & 54.67 & 45.73 \\
    Fast-DetectGPT & 8.02 & 27.46 & 44.01 & 31.84 & 44.26 & 37.03 & 39.88 & 33.22 \\
    \midrule
    \rowcolor{lightgray}
    \multicolumn{9}{c}{{\bf \textit{Phi-4}}} \\
    \midrule
    Entropy & 50.79 & 57.71 & 44.76 & 50.86 & 54.62 & 52.45 & 52.24 & 51.92 \\
    Lastde & 25.83 & 65.61 & 78.99 & 66.94 & 78.14 & 66.27 & 77.95 & 65.67 \\
    Lastde++ & 13.40 & 42.14 & 61.89 & 43.13 & 60.90 & 51.73 & 59.78 & 47.57 \\
    Log-Likelihood & 9.39 & 25.97 & 43.53 & 31.28 & 34.01 & 38.76 & 33.63 & 30.94 \\
    LogRank & 11.41 & 28.60 & 44.26 & 32.22 & 35.73 & 39.98 & 36.02 & 32.60 \\
    Detect-LRR & 26.85 & 40.34 & 47.65 & 36.90 & 43.69 & 44.38 & 44.56 & 40.62 \\
    Fast-DetectGPT & 6.08 & 15.61 & 24.42 & 15.05 & 25.34 & 21.08 & 21.67 & 18.47 \\
    \midrule
    \rowcolor{lightgray}
    \multicolumn{9}{c}{{\bf \textit{Qwen-3-4B}}} \\
    \midrule
    Entropy & 70.23 & 80.26 & 71.89 & 75.36 & 79.37 & 73.63 & 82.53 & 76.18 \\
    Lastde & 23.59 & 60.30 & 72.78 & 68.88 & 76.44 & 61.58 & 74.43 & 62.57 \\
    Lastde++ & 9.88 & 30.06 & 46.55 & 42.82 & 54.65 & 41.56 & 52.94 & 39.78 \\
    Log-Likelihood & 1.72 & 5.68 & 11.97 & 9.00 & 10.62 & 17.14 & 8.52 & 9.23 \\
    LogRank & 2.42 & 7.15 & 12.44 & 9.90 & 12.23 & 18.68 & 10.26 & 10.44 \\
    Detect-LRR & 11.67 & 17.30 & 19.39 & 17.96 & 23.16 & 26.64 & 19.88 & 19.43 \\
    Fast-DetectGPT & 2.91 & 5.29 & 8.67 & 8.78 & 15.34 & 8.26 & 11.72 & 8.71 \\
    \bottomrule
    \end{tabular}
    \caption{Full AUROC results on \texttt{Stylo-Literary}, where generators are \colorbox{lightgray}{\textbf{\textit{highlighted}}}.}
    \label{tab:full_liter}
    \end{table*}

\begin{table*}[htbp]
    \centering\small
    \begin{tabular}{lcccccc}
    \midrule
    \diagbox[dir=SE,height=2\line]{\textbf{Detector}}{\textbf{Subdomain}} & \textbf{arXiv} & \textbf{PeerRead} & \textbf{Reddit} & \textbf{WikiHow} & \textbf{Wikipedia} & \textbf{Avg.} \\
    \midrule
    \rowcolor{lightgray}
    \multicolumn{7}{c}{{\bf \textit{BLOOMz}}} \\
    \midrule
    Entropy & 33.95 & 23.66 & 23.82 & 41.90 & 83.33 & 41.33 \\
    Lastde & 98.98 & 98.94 & 96.38 & 55.84 & 90.03 & 88.03 \\
    Lastde++ & 98.56 & 83.96 & 79.51 & 52.57 & 89.98 & 80.91 \\
    Log-Likelihood & 88.71 & 68.97 & 64.56 & 50.05 & 26.44 & 59.75 \\
    LogRank & 94.39 & 84.96 & 80.29 & 51.85 & 35.62 & 69.42 \\
    Detect-LRR & 98.93 & 98.56 & 97.42 & 57.08 & 69.69 & 84.34 \\
    Fast-DetectGPT & 96.04 & 46.73 & 31.23 & 25.41 & 75.66 & 55.01 \\
    \midrule
    \rowcolor{lightgray}
    \multicolumn{7}{c}{{\bf \textit{ChatGPT}}} \\
    \midrule
    Entropy & 61.64 & 4.21 & 10.35 & 7.04 & 48.49 & 26.35 \\
    Lastde & 99.00 & 99.00 & 99.37 & 92.62 & 97.40 & 97.48 \\
    Lastde++ & 99.85 & 99.85 & 99.72 & 94.86 & 99.07 & 98.67 \\
    Log-Likelihood & 89.13 & 99.45 & 98.98 & 96.57 & 84.69 & 93.76 \\
    LogRank & 91.48 & 99.51 & 99.14 & 96.26 & 88.29 & 94.94 \\
    Detect-LRR & 94.08 & 99.29 & 98.96 & 93.02 & 95.32 & 96.13 \\
    Fast-DetectGPT & 99.93 & 99.92 & 99.84 & 96.22 & 99.03 & 98.99 \\
    \midrule
    \rowcolor{lightgray}
    \multicolumn{7}{c}{{\bf \textit{Cohere}}} \\
    \midrule
    Entropy & 64.16 & 14.39 & 28.84 & 6.62 & 45.16 & 31.83 \\
    Lastde & 95.91 & 97.28 & 99.67 & 99.57 & 96.01 & 97.69 \\
    Lastde++ & 98.50 & 95.93 & 98.74 & 99.90 & 98.04 & 98.22 \\
    Log-Likelihood & 85.67 & 97.02 & 96.08 & 99.43 & 87.39 & 93.12 \\
    LogRank & 88.24 & 96.92 & 96.38 & 99.56 & 89.97 & 94.21 \\
    Detect-LRR & 90.59 & 95.28 & 96.13 & 99.37 & 93.04 & 94.88 \\
    Fast-DetectGPT & 99.50 & 97.00 & 99.20 & 99.93 & 98.30 & 98.78 \\
    \midrule
    \rowcolor{lightgray}
    \multicolumn{7}{c}{{\bf \textit{Davinci}}} \\
    \midrule
    Entropy & 78.65 & 19.92 & 17.00 & 23.26 & 61.68 & 40.10 \\
    Lastde & 50.53 & 99.49 & 94.32 & 85.46 & 88.73 & 83.70 \\
    Lastde++ & 41.47 & 99.81 & 98.34 & 87.80 & 94.64 & 84.41 \\
    Log-Likelihood & 21.97 & 99.11 & 95.22 & 84.26 & 63.40 & 72.79 \\
    LogRank & 20.68 & 99.30 & 95.39 & 83.46 & 67.08 & 73.18 \\
    Detect-LRR & 20.29 & 99.25 & 93.77 & 78.75 & 78.10 & 74.03 \\
    Fast-DetectGPT & 44.09 & 99.91 & 98.72 & 89.36 & 94.29 & 85.28 \\
    \bottomrule
    \end{tabular}
    \caption{Full AUROC results on \texttt{M4}, where generators are \colorbox{lightgray}{\textbf{\textit{highlighted}}}.}
    \label{tab:full_m4}
    \end{table*}

\begin{table*}[htbp]
    \centering\small
    \textcolor{blue}{\begin{tabular}{llcccccccc}
    \toprule
    \textbf{Detector} & \textbf{Generator} & \textbf{J.A} & \textbf{C.D} & \textbf{F.D} & \textbf{P.L} & \textbf{B.S} & \textbf{J.S} & \textbf{M.T} & \textbf{Avg.} \\
    \midrule
    \multirow{3}{*}{\centering Roberta} & Llama-3.1-8B & 35.09 & 66.87 & 68.81 & 46.90 & 64.12 & 50.71 & 63.81 & 56.62 \\
     & Phi-4 & 46.23 & 55.77 & 61.65 & 36.71 & 55.15 & 45.10 & 54.50 & 50.73 \\
     & Qwen3-4B & 58.82 & 58.25 & 54.92 & 37.25 & 57.25 & 45.75 & 56.27 & 52.64 \\
    \midrule
    \multirow{3}{*}{\centering DeTeCtive} & Llama-3.1-8B & 68.98 & 65.31 & 61.80 & 66.34 & 71.73 & 64.02 & 63.21 & 65.91 \\
     & Phi-4 & 63.63 & 63.09 & 68.25 & 65.75 & 69.15 & 63.40 & 65.10 & 65.48 \\
     & Qwen3-4B & 82.99 & 77.16 & 77.41 & 71.48 & 75.94 & 74.01 & 70.29 & 75.61 \\
    \bottomrule
    \end{tabular}
    }
    \caption{\textcolor{blue}{Results of training-based detectors on \texttt{Stylo-Literary}.}}
    \label{tab:mgt---tr-ft}
\end{table*}

\begin{table*}[htbp]
    \centering\small
    \textcolor{blue}{\begin{tabular}{lccccc}
    \toprule
    \textbf{Detector} & \textbf{Claude3.7} & \textbf{Claude4} & \textbf{GPT-4o} & \textbf{Qwen2.5-72B} & \textbf{Avg.} \\
    \midrule
    Roberta & 47.38 & 57.33 & 59.40 & 94.99 & 64.78 \\
    DeTeCtive & 43.97 & 45.06 & 47.68 & 60.39 & 49.28 \\
    \bottomrule
    \end{tabular}
    \caption{\textcolor{blue}{Results of training-based detectors on \texttt{Stylo-Blog}.}}}
    
    \label{tab:mgt---tr-pr}
\end{table*}

\begin{table*}[htbp]
    \centering\small
    \textcolor{blue}{\begin{tabular}{llcccccc}
    \toprule
    \textbf{Detector} & \textbf{Generator} & \textbf{arXiv} & \textbf{PeerRead} & \textbf{Reddit} & \textbf{WikiHow} & \textbf{Wikipedia} & \textbf{Avg.} \\
    \midrule
    \multirow{4}{*}{\centering Roberta} & Davinci & 99.33 & 100.00 & 100.00 & 100.00 & 99.98 & 99.86 \\
     & Cohere & 99.92 & 99.99 & 99.95 & 100.00 & 99.77 & 99.93 \\
     & ChatGPT & 100.00 & 100.00 & 100.00 & 100.00 & 99.95 & 99.99 \\
     & BLOOMZ & 99.98 & 100.00 & 100.00 & 100.00 & 100.00 & 100.00 \\
    \midrule
    \multirow{4}{*}{\centering DeTeCtive} & Davinci & 21.04 & 99.56 & 71.45 & 38.15 & 49.12 & 55.86 \\
     & Cohere & 67.04 & 98.64 & 86.20 & 76.87 & 40.97 & 73.94 \\
     & ChatGPT & 89.73 & 99.83 & 80.51 & 39.62 & 56.75 & 73.29 \\
     & BLOOMZ & 70.95 & 96.80 & 87.49 & 49.23 & 91.62 & 79.22 \\
    \bottomrule
    \end{tabular}}

    \caption{\textcolor{blue}{Results of training-based detectors on the English subset of \texttt{M4}.}}
    \label{tab:mgt---tr-m4}

\end{table*}

\onecolumn
\small
\renewcommand{\arraystretch}{1.15}
\setlength{\tabcolsep}{6pt}

\section{Examples of \dataset}
\label{app:examples}

For an intuitive understanding, we show examples of \texttt{Stylo-Blog} and \texttt{Stylo-Literary} with different generators and sub-domains in Tables~\ref{tab:stylo-blog-long} and~\ref{tab:stylo-literary-long}.

\begin{longtable}{@{}l >{\raggedright\arraybackslash}p{0.82\textwidth}@{}}
\caption{Examples of \texttt{Stylo-Blog}. The first 30 tokens used to generate MGTs are \textbf{bolded}.}
\label{tab:stylo-blog-long} \\
\toprule
\textbf{Model} & \textbf{Text} \\
\midrule
\endfirsthead

\caption[]{Examples of \texttt{Stylo-Blog} (continued)} \\
\toprule
\textbf{Model} & \textbf{Text} \\
\midrule
\endhead

\midrule
\multicolumn{2}{r}{Continued on next page} \\
\endfoot

\bottomrule
\endlastfoot

\rowcolor{lightgray}
\multicolumn{2}{c}{{\bf \textit{Human}}} \\
Human & There is so much on my mind these days...  So much that I can't seem to form a clear thought.  I want so badly to move on with my life, but habits and people and I, myself, keep pulling me back.  I know what I need to do, but it's almost like I am afraid to take that step.  I've always done the safe thing, made the safe choices.  I've always done what was expected of me.  And now that I am finally trying to step out on my own, I'm being told that I am running away.  As much as I know that I need to do this, I still have those same childish fears that I've always had.  I love my family, I love my friends, but I use them as a crutch.  I use them as the reasons to stay in my comfort zone.  I use them as my excuses to not step into that shadowy place just beyond my line of sight.  I am not running away from anything, well, maybe some things, but sometimes, running away is the only way to set yourself free.  I've prayed and prayed, begged and pleaded with God to not tell me what I already know.  There is no way to deny that I know where my life is going.  The only way that I can get to where I truly need to be is to walk away from everything that I know, to take that first blind step into my future.  And the only way to do that is to put my complete trust and total faith in God.\\
\midrule

\rowcolor{lightgray}
\multicolumn{2}{c}{{\bf \textit{Machine}}} \\
Qwen2.5-72B & \textbf{There is so much on my mind these days... So much that I can't seem to form a clear thought. I want so badly to} share all these swirling emotions and thoughts, but the words just don't seem to come. It feels like a storm is brewing inside me, with no outlet in sight. The more I try to wrap my head around everything, the more I feel like I’m losing my grip on the present. \textbackslash n\textbackslash nI've been thinking a lot about the people in my life recently. The relationships that have shaped me, the ones that have endured through thick and thin, and the ones that have faded into the background. It's funny, how the people who are closest to you can be the ones you struggle to open up to the most. I find myself holding back, afraid of what will happen if I let it all out. But then, there are moments, like that one in church, where everything bubbles up to the surface and I can't hold it back.\textbackslash n\textbackslash nLast Sunday, I was standing there, singing those familiar worship songs, and it was like I was transported. The words, \textbackslash "I walk by faith, each step by faith; to live by faith, I put my trust in You,\textbackslash " echoed in my heart. And in that moment, I felt it. I looked around, and the faces of the people surrounding me were both familiar and strange. Familiar, because they’ve been a part of my life for so long. Strange, because life has a way of changing us, sometimes so subtly that we don’t even notice until we stop and really look.\textbackslash n\textbackslash nOf the couple hundred people in that room, I only knew about half. Of those, there were the people who have been constants in my life: the adults who have watched me grow from a girl into a woman, and the children who I have watched grow from toddlers to young adults. There were the people who have stood by me through my weakest moments, holding me up when I thought I would fall. And there were those who I have supported, giving them a bit of my strength when they needed it most.\textbackslash n\textbackslash nBut there was one person, one person who I couldn’t talk to, the one person I wasn’t sure if I could leave. And in that moment, the tears started to sting my eyes. I couldn’t reach out, couldn’t communicate what I was feeling. The song ended in fellowship time, and I \\
\midrule
Claude-4 & \textbf{There is so much on my mind these days... So much that I can't seem to form a clear thought. I want so badly to} move on with my life, but I feel like I'm stuck in this weird limbo where nothing makes sense anymore. You know that feeling when you're standing at a crossroads and every direction looks both terrifying and exciting at the same time? That's exactly where I am right now.\textbackslash n\textbackslash nI keep thinking back to all those times when I was so sure I had everything figured out. Remember when I was eighteen and thought I knew exactly how my life would unfold? Ha! If only that naive girl could see me now. She'd probably be shocked at how different things turned out, but maybe she'd also be proud of the woman I've become through all the unexpected twists and turns.\textbackslash n\textbackslash nThe truth is, I'm tired of feeling like I'm waiting for permission to live my own life. Does that make sense? It's like I've been holding my breath, waiting for some magical moment when everything will click into place and I'll suddenly know what I'm supposed to do next. But what if that moment never comes? What if this uncertainty is just part of being human, part of growing up, part of learning to trust yourself even when you can't see the whole staircase?\textbackslash n\textbackslash nI was talking to my mom the other day, and she said something that really stuck with me. She said, \textbackslash "Honey, sometimes moving on doesn't mean you have all the answers. Sometimes it just means you're brave enough to take the next step without \\
\midrule
Claude-3.7 &\textbf{There is so much on my mind these days... So much that I can't seem to form a clear thought. I want so badly to} just sit and process everything, but life keeps moving at this relentless pace that doesn't allow for much reflection.\textbackslash n\textbackslash nYou know that feeling when your brain is like a browser with too many tabs open? That's me right now. A hundred thoughts racing, memories surfacing at odd moments, worries about tomorrow pressing in, and in the midst of it all, trying to remember if I picked up milk at the store yesterday.\textbackslash n\textbackslash nI found myself driving home from work yesterday, and I couldn't even tell you what streets I took. My body was on autopilot while my mind wandered through a maze of its own making. It's scary sometimes, how disconnected we can become from the present moment.\textbackslash n\textbackslash nLast night, I was folding laundry – such a mundane task – when suddenly tears were streaming down my face. Not sad tears necessarily, just... release tears. The kind that come when your heart is too full of something unnamed. The kind that need no explanation or justification. They just are.\textbackslash n\textbackslash nAnd in that moment, with a half-folded t-shirt in my hands, I realized how much I've been holding in. How many emotions I've been packing away neatly like these clothes, trying to make everything fit in the limited space I've allowed myself.\textbackslash n\textbackslash nI think we all do this sometimes. We keep pushing forward, keep smiling, keep showing up for the people who need us. We sing the worship songs on Sunday morning, we make the lunches, we answer the emails, we check the boxes. And somewhere in all of that, our own hearts get a little lost.\textbackslash n\textbackslash nI'm looking at my Bible sitting on the nightstand. It's been there for days, untouched. Not because I don't want to open it, but because I'm afraid of what might happen if I do. If I get still enough to listen, what will God say? What will He ask of me? What emotions might surface that I've been working so hard to keep submerged?\textbackslash n\textbackslash nI walk by faith, each step by faith.\textbackslash n\textbackslash nThose words from the song at church keep coming back to me. Walking by faith doesn't always mean having clear direction. Sometimes it means putting one foot in front of the other when the path ahead is completely obscured. Sometimes it means trusting that even in the mental chaos, God \\
\midrule
GPT-4o & \textbf{There is so much on my mind these days... So much that I can't seem to form a clear thought. I want so badly to} put it all into words, but every time I try, they swirl around like leaves caught in an autumn breeze. It's as if the very alphabet refuses to settle on the page, preferring instead to dance in chaos. And amidst this whirlwind, I find myself yearning for clarity, for a moment of stillness where I can breathe and just... be.\textbackslash n\textbackslash nSometimes I think about how life used to be so simple, yet we couldn't wait to grow up. Oh, how naive we were! The dreams of adulthood seemed wrapped in wonder and possibility, but now I see that they come twined with responsibility and introspection. I pause, caught in the space between what was and what is, longing for that childlike clarity once more.\textbackslash n\textbackslash nThis introspection takes me back to those quiet moments in the pews of the old church, where the weight of expectations lifted slightly with each hymn sung. There, in the company of familiar faces, I found something akin to peace. We were together, bound not just by belief but by the shared journey of faith and vice-versa, bearing the burdens of our collective humanity.\textbackslash n\textbackslash nYet these days, everywhere I turn, there seems to be complexity and noise. The world spins faster, and sometimes I feel like an outsider peering in through fogged glass. I find solace in small acts: the laughter of a child, the rustle of leaves, the comforting ordinariness of a daily routine. I seek meaning in the minutiae, an anchor in the intricate tapestry of life.\textbackslash n\textbackslash nI've been volunteering more with the local youth group, hoping to give back some of what was freely given to me. And, oh, how their perspectives refresh my own! Their unfiltered view of the world serves as a reminder of what truly matters. I watch them, these bundles of potential and enthusiasm, and think about possibilities — not just their possibilities, but mine too. \textbackslash n\textbackslash nHow often do we, as adults, lose sight of our dreams? They get locked away like treasures in a dusty attic, while we pursue notions of success that may not really belong to us. And yet, every now and then, I'm reminded to dust them off, to recall who I truly am beyond the roles and titles accumulated over the years. It’s like catching a glimpse of your reflection in \\

\end{longtable}
\newpage
\begin{longtable}{@{}l >{\raggedright\arraybackslash}p{0.82\textwidth}@{}}
\caption{Examples of \texttt{Stylo-Literary}. The first 30 tokens used to generate MGTs are \textbf{bolded}.}
\label{tab:stylo-literary-long} \\
\toprule
\textbf{Source} & \textbf{Text} \\
\midrule
\endfirsthead

\multicolumn{2}{c}{Examples of \texttt{Stylo-Literary}, continued from last page.} \\
\toprule
\textbf{Source} & \textbf{Text} \\
\midrule
\endhead

\bottomrule
\endfoot


\rowcolor{lightgray}
\multicolumn{2}{c}{{\bf \textit{Human Author: Mark Twain \quad Generator: Llama-3.1-8B}}} \\
Human & \textbf{“Say, wouldn't it 'a' been nuts if he'd a-been here last night ” “Oh, by George, but wouldn't} it” said Ferguson “Then we'd have seen scientific work Intellect--just pure intellect--away up on the upper levels, dontchuknow Archy is all right, and it don't become anybody to belittle him, I can tell you But his gift is only just eyesight, sharp as an owl's, as near as I can make it out just a grand natural animal talent, no more, no less, and prime as far as it goes, but no intellect in it, and for awfulness and marvelousness no more to be compared to what this man does than--than--Why, let me tell you what he'd have done He'd have stepped over to Hogan's and glanced--just glanced, that's all--at the premises, and that's enough See everything Yes, sir, to the last little detail; and he'll know more about that place than the Hogans would know in seven years Next, he would sit down on the bunk, just as ca'm, and say to Mrs Hogan--Say, Ham, consider that you are Mrs Hogan I'll ask the questions; you answer them ” “All right; go on ” “'Madam, if you please--attention--do not let your mind wander Now, then--sex of the child ' “'Female, your Honor ' “'Um--female Very good, very good Age ' “'Turned six, your Honor ' “'Um--young, weak--two miles Weariness will overtake it then It will sink down and sleep We shall find it two miles away, or less Teeth ' “'Five, your Honor, and one a-coming ' “'Very good, very good, very good, indeed ' You see, boys, he knows a clue when he sees it, when it wouldn't mean a dern thing to anybody else 'Stockings, madam Shoes ' “'Yes, your Honor--both ' “'Yarn, perhaps Morocco ' “'Yarn, your Honor And kip ' “'Um--kip This complicates the matter However, let it go--we shall manage Religion ' “'Catholic, your Honor ' “'Very good Snip me a bit from the bed blanket, please Ah, thanks \\
\midrule
Machine & \textbf{“Say, wouldn't it 'a' been nuts if he'd a-been here last night ” “Oh, by George, but wouldn't} that be too many, though I like to talk it out once, just for pastime; and sometimes, when I'm alone in the shop, I try to make it all come again in dialog, and then it seems as if it had, and I dream about it nights You can imagine yourn over again, same as I do When I get tired of myn, I take up Huck's ” “Why, it was the most--” “Yes--incoherency, that's what it was They just rambled around from one thing to another, and never landed in the same place twice Let it alone; it ain't anybody's fault Consider them illiterates Now you shell notice, Astronomical would naturally drop into their own language; and he did Then he would always try to correct himself, and begin to sling off parts of words that he thought belonged to English, but didn't No, he couldn't put together so you could remember anything he'd said, within fifteen minutes after it was done; and he couldn't get his sentence started twice, right straight through So somebody else would interrupt and say, “What's de use, let him 'lone--dey can't git shet o' de subject,” or something like that, you know, and then start himself and go to talking You see, he tried to talk English all the time, only he couldn't somehow It got him down So he took to talking dialect, because it was easier and convenient, you know Well, as soon as he'd break loose, there warn't no getting him to his english again He couldn't seem to settle down to it We all used to go for him and rush him every time he'd begin to talk English, and force him to talk dialect The Widow Douglas used to break in on me dreadful at first, with her English; but she got the hang of it a deal sooner than any of the rest She said she couldn't stand it, and she warned me off from trying to make a body talk dialect that hadn't been brought up to it, I might as well atrophy an elephant Yes, even the widow: everybody was down to the other fellow's level, before we got through And some better'n others, too Present\\
\addlinespace[4pt]
\midrule
\rowcolor{lightgray}
\multicolumn{2}{c}{{\bf \textit{Human Author: Jane Austen \quad Generator: Llama-3.1-8B}}} \\
Human & \textbf{” So far her improvement was sufficient--and in many other points she came on exceedingly well; for though she could not write sonnets, she brought herself to read them}; and though there seemed no chance of her throwing a whole party into raptures by a prelude on the pianoforte, of her own composition, she could listen to other people's performance with very little fatigue Her greatest deficiency was in the pencil--she had no notion of drawing--not enough even to attempt a sketch of her lover's profile, that she might be detected in the design There she fell miserably short of the true heroic height At present she did not know her own poverty, for she had no lover to portray She had reached the age of seventeen, without having seen one amiable youth who could call forth her sensibility, without having inspired one real passion, and without having excited even any admiration but what was very moderate and very transient This was strange indeed But strange things may be generally accounted for if their cause be fairly searched out There was not one lord in the neighbourhood; no--not even a baronet There was not one family among their acquaintance who had reared and supported a boy accidentally found at their door--not one young man whose origin was unknown Her father had no ward, and the squire of the parish no children But when a young lady is to be a heroine, the perverseness of forty surrounding families cannot prevent her Something must and will happen to throw a hero in her way Mr Allen, who owned the chief of the property about Fullerton, the village in Wiltshire where the Morlands lived, was ordered to Bath for the benefit of a gouty constitution--and his lady, a good-humoured woman, fond of Miss Morland, and probably aware that if adventures will not befall a young lady in her own village, she must seek them abroad, invited her to go with them Mr and Mrs Morland were all compliance, and Catherine all happiness \\
\midrule
Machine & \textbf{” So far her improvement was sufficient--and in many other points she came on exceedingly well; for though she could not write sonnets, she brought herself to read them} with feeling, and the history of her first setting out to excel was soon written, and made into a narrow little volume, two volumes in quinto, and all in her own hand She had no notice of it herself, till her mother one day said, “Ellen, may I see your drawing-room window ” And when Elizabeth turned her hand out of the draw- ing-book, some elegant little figure of a tower or grotto which Miss Crawford had criticising, there, without thinking, lay the pretty little volume The look which Elizabeth instantly raised to comprehend the business, remained some time on Miss Crawford's face, and hers was AN UNPLEASANT ONE It spoke meekness, submitting, shy distrust, penetrating curiosity, but yet grudging invitation A second or two passed in complete silence before either of them spoke, and then it was only, “Well, Ellen --I thought you had been writing ” “Oh I did, indeed, I was so much concerned about it ” “And have you explained everything to Mr Morland What does he say ” “He is very good-aided But I have been used to have poorer friends My family have been poor, for years ” “Yes; that is, you have no father or brother living I beg your pardon You are on the spot yourself, which is always something in a friend He must be a singular man--a silent, melancholy, earnest, persevering, thinking, practicing man--his mind completely under his own control--somewhat severe, perhaps--but with great energy--great forbearance--great capacity of enjoying the thing that he likes --provided a woman's wishes were consulted but what do I tell you They never ARE At this rate you will never know Badly as we judge alone, we should never dare to judge till \\
\addlinespace[4pt]
\midrule
\rowcolor{lightgray}
\multicolumn{2}{c}{{\bf \textit{Human Author: Jane Austen \quad Generator: Phi-4}}} \\
Human & \textbf{” “I am very sorry for it,” said Catherine dejectedly; “if I had known this before--” “Perhaps,” said Miss Tilney} in an embarrassed manner, “you would be so good--it would make me very happy if--” The entrance of her father put a stop to the civility, which Catherine was beginning to hope might introduce a desire of their corresponding After addressing her with his usual politeness, he turned to his daughter and said, “Well, Eleanor, may I congratulate you on being successful in your application to your fair friend ” “I was just beginning to make the request, sir, as you came in ” “Well, proceed by all means I know how much your heart is in it My daughter, Miss Morland,” he continued, without leaving his daughter time to speak, “has been forming a very bold wish We leave Bath, as she has perhaps told you, on Saturday se'nnight A letter from my steward tells me that my presence is wanted at home; and being disappointed in my hope of seeing the Marquis of Longtown and General Courteney here, some of my very old friends, there is nothing to detain me longer in Bath And could we carry our selfish point with you, we should leave it without a single regret Can you, in short, be prevailed on to quit this scene of public triumph and oblige your friend Eleanor with your company in Gloucestershire I am almost ashamed to make the request, though its presumption would certainly appear greater to every creature in Bath than yourself Modesty such as yours--but not for the world would I pain it by open praise If you can be induced to honour us with a visit, you will make us happy beyond expression 'Tis true, we can offer you nothing like the gaieties of this lively place; we can tempt you neither by amusement nor splendour, for our mode of living, as you see, is plain and unpretending; yet no endeavours shall be wanting on our side to make Northanger Abbey not wholly disagreeable ” Northanger Abbey These were thrilling words, and wound up Catherine's feelings to the highest point of ecstasy Her grateful and gratified heart could hardly restrain its expressions within the language of tolerable calmness To receive so flattering an invitation To have her company so warmly solicited \\
\midrule
Machine & \textbf{” “I am very sorry for it,” said Catherine dejectedly; “if I had known this before--” “Perhaps,” said Miss Tilney}, with an affectionate smile, “it would not have made you more easy on the subject ” Her brother instantly interfered; “Do you advise her to tell me of it then ” “By no means, if she can help it The disadvantage is at least mutual There need be no consequence but what will be attended by pleasure ” But though the matter was apparently arranged in a moment, and they all three seemed most happily reconciled to each other, there yet lingered in Catherine’s mind some feelings which required the subduing influence of a good night’s rest to put them quite to flight; and, though without exactly knowing why, as soon as she left the room, and while slowly ascending the staircase, she felt convinced that General Tilney could never like her, or think her pretty --They had one family friend, besides these gentlemen, who resided among them Mr Frederick Tilney, their eldest son, was the rector of the parish, of which his father was patron A young man very like his cousin John, but handsomer than him He had a great deal of the same manner--and every body thought him extremely agreeable, except Catherine She did not like him; and, now that Edmund Bertram was removed from her acquaintance, she wished very much that \_he\_ might never return from Oxford It appeared afterwards that he must often visit his relations in Gloucestershire, and this raised a thousand unpleasant questions: Whom would he come with Why was not his sister invited also  Would he bring Anne Heretitia Mention of Mr Allen did so little honour to either sex, to human nature in general, that Catherine was rather glad the circumstance was suppressed They sat down to table, twenty-three in number, and in a style of luxury, which, to Catherine, with the exception of her first morning’s entertainment, was peculiar Having been indulged with uncommon orders in her favourite toy-shop only two days before, she knew how much superior was cold beef, or cold lamb, at any time of life, to the daintiest dinner that ever was cooked; and being prepared, therefore, to find fault, nothing could be more exquisite than her admiration In addition to every other \\
\addlinespace[4pt]
\midrule
\rowcolor{lightgray}
\multicolumn{2}{c}{{\bf \textit{Human Author: Bernard Shaw \quad Generator: Phi-4}}} \\
Human & \textbf{We shall come to the question of the economic dependence of women on men later on; but at present we had better confine ourselves to the theories of} marriage which we are not ashamed to acknowledge and defend, and upon which, therefore, marriage reformers will be obliged to proceed We may, I think, dismiss from the field of practical politics the extreme sacerdotal view of marriage as a sacred and indissoluble covenant, because though reinforced by unhappy marriages as all fanaticisms are reinforced by human sacrifices, it has been reduced to a private and socially inoperative eccentricity by the introduction of civil marriage and divorce Theoretically, our civilly married couples are to a Catholic as unmarried couples are: that is, they are living in open sin Practically, civilly married couples are received in society, by Catholics and everyone else, precisely as sacramentally married couples are; and so are people who have divorced their wives or husbands and married again And yet marriage is enforced by public opinion with such ferocity that the least suggestion of laxity in its support is fatal to even the highest and strongest reputations, although laxity of conduct is winked at with grinning indulgence; so that we find the austere Shelley denounced as a fiend in human form, whilst Nelson, who openly left his wife and formed a menage a trois with Sir William and Lady Hamilton, was idolized Shelley might have had an illegitimate child in every county in England if he had done so frankly as a sinner His unpardonable offence was that he attacked marriage as an institution We feel a strange anguish of terror and hatred against him, as against one who threatens us with a mortal injury What is the element in his proposals that produces this effect The answer of the specialists is the one already alluded to: that the attack on marriage is an attack on property; so that Shelley was something more hateful to a husband than a horse thief: to wit, a wife thief, and something more hateful to a wife than a burglar: namely, one who would steal her husband's house from over her head, and leave her destitute and nameless on the streets\\
\midrule
Machine & \textbf{We shall come to the question of the economic dependence of women on men later on; but at present we had better confine ourselves to the theories of} Mrs Warren's profession As far as its religious aspect goes, it seems, so far as I can make out, to be a practice of prostitution in which no payment is accepted and from which the practitioner expects nothing but damnation This has never been a very popular method: St Paul tells us that prostitutes were always ready enough to trade for money without asking for any recompense in heaven Consequently it was not until Mr Gladstone found out, when he visited Florence in 1897, that Fra Angelico represented angels with wings instead of tails, that a reaction set in against Dante, who did not take this view of things There are many saints in our calendar, male and female, whose sanctity has apparently depended largely upon their experience of this form of the calling But these records are extremely obscure, because they all insist much more strongly than Dante does on the fact that there is only one thing wrong with vice: that is, that it should pay Now in all discussions as to whether there is anything fundamentally wrong with human nature--the strongest practical grounds being generally alleged by those who believe nothing whatever in its fundamental soundness--it is necessary to have a common denominator Both sides must agree what \"wrong\" means before they can go on to argue about the degree in which it exists If there is something really wrong with human nature, then, logically, no human conduct or instinct or desire can possibly be regarded as infallibly right The immorality of an action cannot depend entirely on the motive behind it You may begin by admitting that if you do certain things with a good motive your act is not sinful; but still, you cannot admit that such acts done with a bad motive are not wicked Why should not the most disinterested and self-sacrificing promiscuity be just as vicious as the greediest and vilest lust And why should not the vilest and greediest lust be just as virtuous as the purest devotion Can anyone answer these questions sensibly (I am sure nobody ever tries) except on the\\
\addlinespace[4pt]
\midrule
\rowcolor{lightgray}
\multicolumn{2}{c}{{\bf \textit{Human Author: Bernard Shaw \quad Generator: Qwen3-4B}}} \\
Human &\textbf{ We shall come to the question of the economic dependence of women on men later on; but at present we had better confine ourselves to the theories of} marriage which we are not ashamed to acknowledge and defend, and upon which, therefore, marriage reformers will be obliged to proceed We may, I think, dismiss from the field of practical politics the extreme sacerdotal view of marriage as a sacred and indissoluble covenant, because though reinforced by unhappy marriages as all fanaticisms are reinforced by human sacrifices, it has been reduced to a private and socially inoperative eccentricity by the introduction of civil marriage and divorce Theoretically, our civilly married couples are to a Catholic as unmarried couples are: that is, they are living in open sin Practically, civilly married couples are received in society, by Catholics and everyone else, precisely as sacramentally married couples are; and so are people who have divorced their wives or husbands and married again And yet marriage is enforced by public opinion with such ferocity that the least suggestion of laxity in its support is fatal to even the highest and strongest reputations, although laxity of conduct is winked at with grinning indulgence; so that we find the austere Shelley denounced as a fiend in human form, whilst Nelson, who openly left his wife and formed a menage a trois with Sir William and Lady Hamilton, was idolized Shelley might have had an illegitimate child in every county in England if he had done so frankly as a sinner His unpardonable offence was that he attacked marriage as an institution We feel a strange anguish of terror and hatred against him, as against one who threatens us with a mortal injury What is the element in his proposals that produces this effect The answer of the specialists is the one already alluded to: that the attack on marriage is an attack on property; so that Shelley was something more hateful to a husband than a horse thief: to wit, a wife thief, and something more hateful to a wife than a burglar: namely, one who would steal her husband's house from over her head, and leave her destitute and nameless on the streets \\
\midrule
Machine &\textbf{We shall come to the question of the economic dependence of women on men later on; but at present we had better confine ourselves to the theories of} immorality and bad manners which is, in short, one single doctrine which has grown by multiplication into an infinity of articles To all these it may be said that the theory to which they are attributable is the theory of physical inferiority, moral incapacity, inhumanity or mere unskilled destructiveness The elementary creed includes not only the theory of male superiority as actually evidenced in modern warfare, but also such absurd assumptions as that in spite of modern civilized warfare women cannot fight and must therefore remain separate from mankind even if they have no more powerful reason for doing so; or again that the splendid courage displayed by the women who stood out during the Great War was produced by a preternatural instinctive force: in other words, a sexual instinctuality, which is declared to be incapable of being cultivated by education All this is gobbledygook made up by professional female critics to please their paymasters The truth in short is what Adam Smith, long ago put it thus \"To man's nature is due his love of dominion over woman He may, and I believe he ought sometimes to, allow her freedom of choice in marriage matters; but when he does this, he leaves his wife in much the same position as if she were married to him against her will A great deal of what we call justice or respectability between husband and wife really comes from a strong sense of self-reverence, accompanied by an equally strong sense of the infamy of a lie \" THE HOMO SAPIENS MUSICAL PARADISE Some people feel perfectly happy in the belief that women are more refined than men because they are less brutal, less selfish, less rapacious In itself this idea makes some sort of sense; but it has been found that it does not explain everything Every human being desires, sooner or later, some sort of gratification Men usually seek satisfaction from sex activity, and sometimes from sublimated forms of that activity (theology, science, music); whereas women often find no satisfactory substitute except bare animal \\

\end{longtable}

\end{document}